%% file: MAIN.tex
\documentclass[pmlr]{jmlr}

\usepackage{graphicx}
\usepackage{subcaption}
\usepackage{xcolor,colortbl}

\usepackage{footnote}
\usepackage[]{threeparttable}
\makesavenoteenv{tabular}
\usepackage{algorithm}
\usepackage[noend]{algpseudocode}
\usepackage{xr}

\usepackage{longtable}
\usepackage{multirow}

\usepackage{ upgreek }
\usepackage{dsfont}

\usepackage{booktabs}
\usepackage[load-configurations=version-1]{siunitx} 

\usepackage{float}
\makeatletter
\def\set@curr@file#1{\def\@curr@file{#1}} 
\makeatother

\theorembodyfont{\upshape}
\theoremheaderfont{\scshape}
\theorempostheader{:}
\theoremsep{\newline}

\usepackage{shortbold}

\jmlrvolume{126}
\jmlryear{2020}
\jmlrworkshop{Machine Learning for Healthcare}

\title[Exploiting and Preventing Shortcuts]{Deep Learning Applied to Chest X-Rays:\\ Exploiting and Preventing Shortcuts}

\author{\Name{Sarah Jabbour}
\Email{\href{mailto:sjabbour@umich.edu}{\color{black}{sjabbour@umich.edu}}}
\addr {\\ \footnotesize Department of Electrical Engineering and Computer Science \\
University of Michigan, Ann Arbor, MI, USA} 
\AND
\Name{David Fouhey}
\Email{\href{mailto:fouhey@umich.edu}{\color{black}{fouhey@umich.edu}}} 
\addr {\\ \footnotesize Department of Electrical Engineering and Computer Science \\
University of Michigan, Ann Arbor, MI, USA} 
\AND
\Name{Ella Kazerooni}
\Email{\href{mailto:ellakaz@med.umich.edu}{\color{black}{ellakaz@med.umich.edu}}}
\addr {\\ \footnotesize Department of Radiology; and \\
Department of Internal Medicine \\ 
University of Michigan, Ann Arbor, MI, USA} 
\AND
\Name{Michael W. Sjoding}
\Email{\href{mailto:msjoding@med.umich.edu}{\color{black}{msjoding@med.umich.edu}}}
\addr {\\ \footnotesize Institute for Healthcare Policy \& Innovation; and \\
Department of Internal Medicine \\
University of Michigan, Ann Arbor, MI, USA}
\AND
\Name{Jenna Wiens}
\Email{\href{mailto:wiensj@umich.edu}{\color{black}{wiensj@umich.edu}}}
\addr {\\ \footnotesize Department of Electrical Engineering and Computer Science \\
University of Michigan, Ann Arbor, MI, USA} 
}

\begin{document}

\maketitle

\input{mlhc-submission-files/text/abstract_into_related_work.tex}

\input{mlhc-submission-files/text/proposed_approach.tex}

\input{mlhc-submission-files/text/experiments.tex}

\input{mlhc-submission-files/text/discussion}

\acks{This work was supported by the National Science Foundation (NSF award no. IIS-1553146) and the National Library of Medicine (NLM grant no. R01LM013325). The views and conclusions in this document are those of the authors and should not be interpreted as necessarily representing the official policies, either expressed or implied, of the National Science Foundation nor the National Library of Medicine. We would like to thank the MIMIC-CXR team for providing attribute labels for their dataset.}

\bibliography{ref}
\clearpage

\appendix
\section{}

\begin{table*}[ht]
    \centering
    \caption{\label{tab:correlations} Spearman's correlation between the 14 tasks in MIMIC-CXR and CheXpert and attributes $b$ (after mapping uncertain labels to positive, all No Finding labels were positive in MIMIC-CXR). We tested for statistical significance with a Bonferroni correction for multiple hypotheses (* representing statistical significance at p-value $< \alpha = 0.001$). With the exceptions of No Finding in MIMIC-CXR and Pneumothorax in both datasets, all tasks were significantly correlated with age. Additionally, many were correlated with sex, insurance, or marital status in either dataset.}
\resizebox{\textwidth}{!}{%
\begin{tabular}{c c c c c c c}
\toprule
 \textbf{Task}& \multicolumn{2}{c}{\textbf{Age}} & \multicolumn{2}{c}{\textbf{Sex}} & \textbf{Insurance} & \textbf{Marital Status} \\
 & \textbf{MIMIC-CXR} & \textbf{CheXpert} &  \textbf{MIMIC-CXR} & \textbf{CheXpert} & \textbf{MIMIC-CXR} & \textbf{MIMIC-CXR} \\
\midrule
 \textbf{No Finding} & -- & $-0.06^*$ & -- & $-0.01$ & -- & -- \\
   \textbf{Enlarged Cardiomediastinum} & $0.20^*$ & $0.20^*$ & $-0.02$ & $-0.03^*$ & $-0.07^*$ & $0.02$ \\
 \textbf{Cardiomegaly} & $0.22^*$ & $0.19^*$ & $0.01$ & $0.00$ & $-0.06^*$ & $0.00$ \\
 \textbf{Airspace Opacity} & $0.05^*$ & $0.08^*$ & $-0.01$ & $0.00$ & $0.00$ & $0.01$ \\
 \textbf{Lung Lesion} & $0.07^*$ & $0.08^*$ & $-0.02$ & $0.02$ & $-0.02$ & $-0.01$ \\
 \textbf{Edema} & $0.18^*$ & $0.08^*$ & $0.00$ & $0.03^*$ & $-0.02$ & $-0.02$ \\
 \textbf{Consolidation} & $0.09^*$ & $0.13^*$ & $-0.07^*$ & $-0.01$ & $-0.04^*$ & $0.04^*$ \\
 \textbf{Pneumonia} & $0.10^*$ & $0.04^*$ & $-0.06^*$ & $0.00$ & $0.00$ & $-0.01$ \\
 \textbf{Atelectasis} & $0.05^*$ & $0.04^*$ & $0.00$ & $-0.01$ & $-0.01$ & $0.01$ \\
 \textbf{Pneumothorax} & $0.00$ & $0.01$ & $-0.03^*$ & $-0.01$ & $-0.01$ & $0.02$ \\
 \textbf{Pleural Effusion} & $0.24^*$ & $0.18^*$ & $-0.02^*$ & $0.01$ & $-0.07^*$ & $0.02^*$ \\
 \textbf{Pleural Other} & $0.12^*$ & $0.09^*$ & $-0.06$ & $-0.04$ & $-0.05$ & $-0.02$ \\
 \textbf{Fracture} & $0.09^*$ & $0.16^*$ & $-0.04$ & $-0.04^*$ & $-0.05$ & $0.01$ \\
 \textbf{Support Devices} & $0.06^*$ & $0.02^*$ & $-0.01$ & $0.02^*$ & $0.01$ & $0.00$ \\
 \bottomrule
\end{tabular}
}
\end{table*}

\begin{table*}[ht]
    \centering
    \caption{\label{tab:skewed_unskewed_datasets} We resampled the unskewed training set such that it was comparable in size to the skewed training set. To mimic the test set distribution, there was no correlation between $y_t$ and each attribute in the unskewed training set. On the other hand, there was a 1:1 correlation between $y_t$ and each attribute in the skewed training set.}
\resizebox{\textwidth}{!}{%
\begin{tabular}{c c c c c c}
\toprule
\textbf{Dataset}  & \textbf{Attribute} & \multicolumn{2}{c}{\textbf{Unskewed Training Set}} & \multicolumn{2}{c}{\textbf{Skewed Training Set}} \\  & & \textbf{N Train (\%)} & \textbf{N Valid (\%)} &  \textbf{N Train (\%)} & \textbf{N Valid (\%)}  \\ 
 \midrule
\multirow[c]{5}{*}{\textbf{AHRF}} & \textbf{Age} &  $508$ $(0.21)$ & $129$ $(0.26)$ &$528$ $(0.26)$ & $139$ $(0.19)$\\
& \textbf{Sex} &  $457$ $(0.10)$ & $114$ $(0.10)$ & $479$ $(0.13)$ & $127$ $(0.10)$
 \\
& \textbf{BMI} &  $477$ $(0.10)$ & $119$ $(0.12)$ & $488$ $(0.12)$ & $127$ $(0.13)$
\\
& \textbf{Race} & $644$ $(0.02)$ & $157$ $(0.03)$ & $679$ $(0.01)$ & $177$ $(0.02)$
 \\ 
& \textbf{Pacemaker} &  $785$ $(0.06)$ & $192$ $(0.07)$ & $757$ $(0.05)$ & $195$ $(0.06)$
 \\ 

 \midrule
\multirow[c]{4}{*}{\textbf{MIMIC-CXR}} & \textbf{Age} & 14225 (0.57) & 1000 (0.55) & 16458 (0.46) & 1000 (0.47) \\ 
& \textbf{Sex} & 14859 (0.52) & 1000 (0.52) & 16641 (0.46) & 1000 (0.48) \\ 
& \textbf{Marital} & 9573 (0.45) & 1000 (0.46) & 10891 (0.47) & 1000 (0.43) \\ 
& \textbf{Insurance} & 1009 (0.53) & 1000 (0.51) & 1201 (0.46) & 1000 (0.49) \\  

 \midrule
 
 \multirow[c]{2}{*}{\textbf{CheXpert}} & \textbf{Age} & 7060 (0.89) & 1000 (0.87) & 7835 (0.90) & 1000 (0.89) \\ 
& \textbf{Sex} & 5702 (0.84) & 1000 (0.85) & 6412 (0.90) & 1000 (0.90) \\
\bottomrule
\end{tabular}
}
\end{table*}

\begin{table}[ht]
\centering
 \caption{\label{tab:statistical_tests_us}
 Statistical significance of the improvement in performance of the model trained on an unskewed dataset versus a model trained on a skewed dataset. The model trained on the unskewed dataset performs significantly better for five of the seven attributes.}

\begin{tabular}{cccc}

\toprule
\textbf{Task}  & \multicolumn{3}{c}{\textbf{p-value}}\\ &  \textbf{AHRF} & \textbf{MIMIC-CXR} & \textbf{CheXpert} \\ 
\midrule 
\textbf{Age}  &  0.20 &  0.00 &  0.00  \\ 
\textbf{Sex}  & 0.02 &  0.00 &  0.00  \\ 
\textbf{BMI} & 0.03 & -- & -- \\
\textbf{Race} & 0.00 & -- & -- \\
\textbf{Pacemaker} & 0.70  & -- & -- \\
 \textbf{Insurance} & -- &  0.00 & \\
\textbf{Marital} & -- & 0.00 & --\\

  \bottomrule 
 \hspace{\fill}

\end{tabular}
\end{table}

\begin{table}[!htp]

\centering
\caption{\label{tab:baseline_results_table} Across all attributes, \texttt{Last Layer M/C + A} yields consistently higher AUROCs compared to \texttt{All Layers}. Furthermore, there is a drop in the predictive performance of each attribute $b$ (i.e., AUROC($\hat{y}$, $b$)) compared to \texttt{All Layers} and \texttt{All Layers M/C}, and a further drop in performance compared to \texttt{Last Layer M/C} when $b$=age. This suggests that the proposed transfer learning approach encourages models to rely \textit{less} on shortcut features an more on clinically relevant attributes of CHF. Additionally, the models trained using the AHRF source task dataset (\texttt{Last Layer M/C + A}) show consistent improvements in performance (with the exception of race) over just using MIMIC-CXR and CheXpert during pretraining (\texttt{All Layers M/C} and \texttt{Last Layers M/C}). Somewhat surprisingly, simply applying a classifier to the features of a CNN allows us to predict CHF with good discriminative performance, while also mitigating, in part, the use of shortcuts.}
\resizebox{\textwidth}{!}{%
\begin{tabular}{c c c c c c c c c}
\toprule
\textbf{Approach} & \multicolumn{5}{c}{\textbf{CHF AUROC (95\% CI)}} \\ &  \textbf{Age} & \textbf{Sex} & \textbf{BMI} &  \textbf{Race} & \textbf{Pacemaker}  \\ 
 \midrule
  \textbf{All Layers} & 0.66 (0.54-0.77)
                    & 0.59 (0.46-0.70)
                    & 0.55 (0.41-0.69) 
                    & 0.60 (0.48-0.73) 
                    &  0.55 (0.42-0.67) \\
                    
 \textbf{All Layers M/C} & 0.71 (0.59-0.82) 
                    & 0.71 (0.59-0.82) 
                    & 0.61 (0.49-0.73)
                    & 0.78 (0.67-0.87) 
                    & 0.81 (0.70-0.89 \\
                
\textbf{Last Layer M/C}  & 0.78 (0.67-0.87) 
                                & 0.72 (0.62-0.82) 
                                & 0.74 (0.62-0.84) 
                                & 0.82 (0.73-0.90)
                                & 0.82 (0.73-0.90) \\
                    
\textbf{Last Layer M/C + A}   & 0.84 (0.73-0.92) 
                        & 0.73 (0.62-0.83) 
                        & 0.78 (0.66-0.87)
                        & 0.82 (0.72-0.89) 
                        & 0.86 (0.78-0.92) \\

\bottomrule
\end{tabular}}

\bigskip
\resizebox{\textwidth}{!}{%
\begin{tabular}{c c c c }
\toprule
\textbf{Approach} & \multicolumn{3}{c}{\textbf{AUROC($\hat{y},b$) (95\% CI)}} \\ &  \textbf{Age} & \textbf{Sex} & \textbf{BMI}  \\ 
 \midrule

 \textbf{All Layers} & 0.81 (0.71-0.89)
                    & 0.92 (0.86-0.96)
                    & 0.81 (0.73-0.89) \\
                    
 \textbf{All Layers M/C}& 0.79 (0.69-0.87)
                    & 0.87 (0.79-0.93)
                    & 0.81 (0.72-0.88)\\

\textbf{Last Layer M/C}  & 0.77 (0.67-0.85)
                        & 0.79 (0.70-0.87)
                        & 0.74 (0.65-0.83)\\
                    
\textbf{Last Layer M/C + A}  & 0.73 (0.63-0.83)
                        & 0.84 (0.76-0.92)
                        & 0.75 (0.66-0.83) \\
\bottomrule
\end{tabular}}

\end{table}

\begin{table*}[ht]
    \centering
    \caption{\label{tab:statistical_tests} Statistical significance of the improvement in performance of \texttt{Last Layer M/C + A} over \texttt{All Layers}. \texttt{Last Layer M/C + A} performs significant better over \texttt{All Layers} for all attributes. }
\scalebox{0.85}{
\begin{tabular}{ c c}
\toprule
& \textbf{p-value} \\ 
 \midrule
\textbf{Age} &  0.00 \\
\textbf{Sex} &  0.00 \\
\textbf{BMI} &  0.00 \\
\textbf{Race} &  0.00 \\
\textbf{Pacemaker} &  0.00 \\
 \bottomrule
\end{tabular}
}
\end{table*}

\begin{table*}[ht]
    \centering
    \caption{
\label{tab:tests_gauss_us}Statistical significance of the improvement in performance of the model trained on an unskewed dataset versus a model trained on a skewed dataset. The model trained on the unskewed dataset performs significantly better for two of the three tasks.}
\scalebox{0.85}{
\begin{tabular}{ c c}
\toprule
& \textbf{p-value} \\ 
 \midrule
\textbf{difficult shortcut} &  0.23 \\
\textbf{moderate shortcut} &  0.00\\
\textbf{easy shortcut} & 0.01 \\
 \bottomrule
\end{tabular}
}
\end{table*}

\begin{table*}[!htpb]
    \centering
    \caption{\label{tab:Gaussian_filter_baselines} \texttt{Last Layer M/C + A} outperforms \texttt{All Layers} on all 3 synthetic datasets, regardless of how difficult the shortcut is to learn. Moreover, \texttt{Last Layer M/C + A} performs consistently worse in terms of the predictive performance of $b$ (AUROC($\hat{y}$, $b$)). This suggests that the proposed transfer learning approach encourages models to rely \textit{less} on shortcut features an more on clinically relevant attributes of CHF. Additionally, \texttt{Last Layer M/C + A} does not outperform training on an unskewed dataset when the shortcuts are easier to learn, i.e. \textit{moderate shortcut} and \textit{easy shortcut}. However, when the shortcut is more difficult to learn, \texttt{Last Layer M/C + A} performs just as well as training on an unskewed dataset (AUROC=0.82 [95\% CI: 0.75-0.87] vs 0.82 [95\% CI: 0.76-0.88]). Finally, \texttt{Last Layer M/C + A} performs better than both \texttt{All Layers M/C} and \texttt{Last Layer M/C}, suggesting that retuning all layers on the AHRF source task incorporates more domain-specific knowledge into pretraining, before training on the biased target task.}

\resizebox{\textwidth}{!}{%
\begin{tabular}{c c c c }
\toprule
\textbf{Approach} & \multicolumn{3}{c}{\textbf{CHF AUROC (95\% CI)}} \\ & \textbf{difficult shortcut}& \textbf{moderate shortcut}   & \textbf{easy shortcut}\\ 
 \midrule

\textbf{All Layers} 
                    &  0.62 (0.54 - 0.70)
                    & 0.51 (0.43 - 0.60)
                    & 0.53 (0.45 - 0.62) \\
                    
\textbf{All Layers M/C} 
                    &  0.83 (0.77 - 0.89)
                    & 0.64 (0.55 - 0.73)
                    & 0.59 (0.51 - 0.67) \\

\textbf{Last Layer M/C} 
                    & 0.78 (0.71 - 0.84)
                    & 0.68 (0.59 - 0.76)
                    & 0.65 (0.58 - 0.73) \\
                    
\textbf{Last Layer M/C + A}  
                    & 0.82 (0.75 - 0.87)
                    & 0.74 (0.66 - 0.81) 
                    & 0.69 (0.61 - 0.76) \\
\midrule
\textbf{Unskewed}  &  \multicolumn{3}{c}{--------------------------- 0.82 (0.76 - 0.88)--------------------------- }\\

 \bottomrule
\end{tabular}}

\bigskip
\resizebox{\textwidth}{!}{%
\begin{tabular}{c c c c }
\toprule
\textbf{Approach} & \multicolumn{3}{c}{\textbf{AUROC($\hat{y},b$) (95\% CI)}} \\ &  \textbf{difficult shortcut} & \textbf{moderate shortcut} & \textbf{easy shortcut}  \\ 
 \midrule

 \textbf{All Layers} & 0.82 (0.75-0.87)
                    & 0.74 (0.66-0.81)
                    & 0.69 (0.61-0.76) \\
                    
 \textbf{All Layers M/C}& 0.81 (0.75-0.87)
                    & 0.66 (0.57-0.74)
                    & 0.64 (0.55-0.71)\\

\textbf{Last Layer M/C}  & 0.73 (0.65-0.79)
                        & 0.63 (0.53-0.72)
                        & 0.54 (0.45-0.62)\\
                    
\textbf{Last Layer M/C + A}  & 0.67 (0.59-0.75)
                        & 0.47 (0.37-0.55)
                        & 0.54 (0.46-0.62) \\
\bottomrule
\end{tabular}}

\end{table*}

\begin{table*}[ht]
    \centering
    \caption{\label{tab:statistical_tests_synthetic} Statistical significance of the improvement in performance of \texttt{All Layers} vs. \texttt{Last Layer M/C + A} on the synthetic datasets. The transfer learning approach performs significantly better for the \textit{easy shortcut}.}
\scalebox{0.85}{
\begin{tabular}{ c c}
\toprule
& \textbf{p-value} \\ 
 \midrule
\textbf{difficult shortcut} & 0.77 \\
\textbf{moderate shortcut} & 0.24 \\
\textbf{easy shortcut} & 0.00 \\
 \bottomrule
\end{tabular}
}
\end{table*}

\clearpage

\subsection{Sensitivity Analysis}
\label{sensitivity analysis}

We performed a sensitivity analysis of \texttt{Last Layer M/C + A} in which, after pretraining on MIMIC-CXR and CheXpert and then the unbiased source task $y_s$, we tuned more layers of the network (instead of just the last layer) when learning to predict $y_t$. If $y_t$ is spuriously correlated with some attribute $b$, we hypothesize that tuning more layers on the target task will allow the model to learn shortcut features related to $b$, leading to poor generalization performance on the test set in which $y_t$ is not correlated with $b$. We use the same DenseNet-121 architecture as presented in \textbf{Section \ref{architecture}}, and vary the number of denseblocks tuned from the end of the network to the beginning, increasing from 0 denseblocks (i.e., \texttt{Last Layer M/C + A}) to all denseblocks (i.e., training the whole network).
\\
\par
\begin{figure}[!ht]
    \centering
    \resizebox{\textwidth}{!}{%
    \includegraphics[scale=0.3]{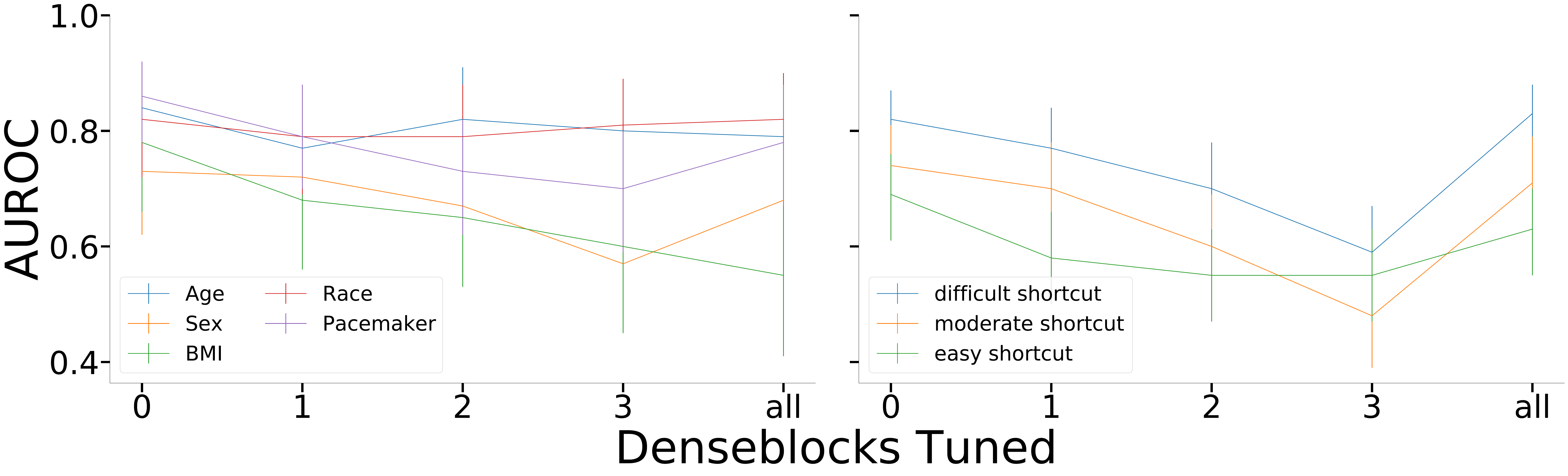}
    }
    \caption{Model performance decreases as the number of trained denseblocks increases. However, model performance increases for some attributes $b$ when we tune all layers of the network compared to tuning three denseblocks. We suspect that, as the parameter space increases from tuning three denseblocks to tuning all layers, the model might need longer to train.}
    \label{fig:sensitivity}
\end{figure}

\begin{figure}[!ht]
    \centering
    \resizebox{\textwidth}{!}{%
    \includegraphics[scale=0.3]{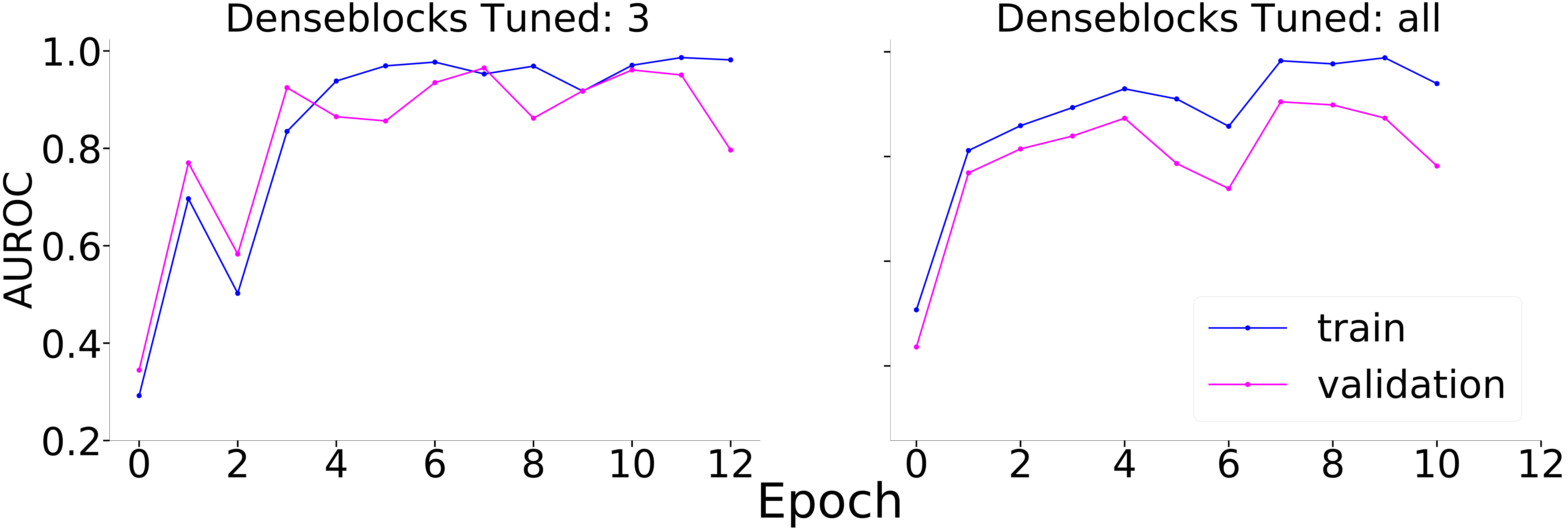}
    }
    \caption{When comparing the training and validation AUROC performance of training three denseblocks to tuning all layers for \textit{moderate shortcut}, training performance for tuning all layers takes longer to fit to the training data. Moreover, there is a bigger gap between training and validation performance when tuning all layers compared to tuning three denseblocks. This suggests that training all layers for \textit{longer} might encourage the model to fit to the data better, and thus learn more shortcuts, leading to a decrease in generalization performance on the unbiased test set.}
    \label{fig:training_curves}
\end{figure}

 \noindent \textbf{Results and Discussion.} As expected, model performance decreases as the number of tuned denseblocks increases (\textbf{Figure \ref{fig:sensitivity}}). However, model performance increases for some attributes $b$ when we tune all layers of the network compared to tuning three denseblocks. We suspect that, as the parameter space increases from tuning three denseblocks to tuning all layers, the model might need longer to train. For example, we compared the training and validation AUROC performance between training three denseblocks and tuning all layers for \textit{moderate shortcut} (\textbf{Figure \ref{fig:training_curves}}). Compared to tuning three denseblocks, training performance for tuning all layers takes longer to fit to the training data. Moreover, there is a bigger gap between training and validation performance when tuning all layers compared to tuning three denseblocks, suggesting that training the model for longer might encourage the model to fit to the data better.

\clearpage

\subsection{Multitask Approach}
\label{multitask} 
We considered an approach based on multitask learning in which a model is trained to jointly predict $y_s$ and $y_t$, minimizing the sum of the respective cross-entropy losses. We hypothesize that jointly training a model to predict $y_s$ and $y_t$ may mitigate the use of shortcut features and perform similarly to the transfer learning approach. Since $\hat{y_s}$ and $\hat{y_t}$ share a feature space, the model will be forced to learn features that are beneficial in predicting \textit{both} outcomes. And, since $y_s$ is not correlated with attribute $b$, the model cannot use shortcuts in predicting $y_s$. These `clinically relevant' features can then also be used in predicting $y_t$. However, we note that the model may benefit from learning some features that are relevant for $y_s$ and others that are relevant for $y_t$, i.e., those features that are relevant for $b$. Additionally, this approach requires labels $y_t$ \textit{and} $y_s$ at training time, which is not always possible. In contrast, the presented transfer learning approach does not rely on access to the training data for the source task, but only the associated model.  In this alternate approach, we use the same DenseNet-121 architecture as presented in \textbf{Section \ref{architecture}}. 
\\
\par \noindent \textbf{Results and Discussion.}  Across five out of eight attributes, \texttt{Last Layer M/C + A} outperforms \texttt{Multitask}. This suggests that the multitask approach, although sharing features between the unbiased source task $y_s$ and biased target task $y_t$, still allows the model to learn shortcut features (\textbf{Table \ref{tab:multitask_results_table}}).

\begin{table}[!ht]

\centering
\caption{\label{tab:multitask_results_table} Across five out of eight attributes, \texttt{Last Layer M/C + A} outperforms the multitask approach. This suggests that the multitask approach, although sharing features between the unbiased source task $y_s$ and biased target task $y_t$, allows the model to learn shortcut features.}
\begin{tabular}{c c c c c c c c c}
\toprule
\textbf{Task} & \multicolumn{2}{c}{\textbf{CHF AUROC (95\% CI)}} \\ &  \textbf{Last Layer M/C + A} & \textbf{Multitask} \\ 
 \midrule
 
 \textbf{Age}  & 0.84 (0.73-0.92)  & 0.76 (0.64-0.86)\\
 \textbf{Sex} & 0.73 (0.62-0.83)   & 0.65 (0.52-0.77)\\
 \textbf{BMI}   & 0.78 (0.66-0.87)  & 0.70 (0.59-0.81)\\
 \textbf{Race}  & 0.82 (0.72-0.89)   & 0.82 (0.74-0.90)\\
 \textbf{Pacemaker} & 0.86 (0.78-0.92)  & 0.87 (0.79-0.93) \\ 
 \textbf{difficult shortcut} & 0.82 (0.75-0.87)  & 0.84 (0.79-0.89) \\ 
 \textbf{moderate shortcut} & 0.74 (0.66-0.81)  & 0.69 (0.60-0.78) \\ 
 \textbf{easy shortcut} & 0.69 (0.61-0.76)  & 0.58 (0.50-0.67) \\ 
                    
\bottomrule
\end{tabular}
\end{table}

\clearpage

\subsection{Gaussian Filter Tasks}

We applied a Gaussian filter with standard deviation $s_1$ to images with $b = 1$ and standard deviation $s_2$ to images with $b = 0$. By varying the strength of the filters, we changed how easily the model could learn shortcuts related each task. We defined the \textit{difficult shortcut} as $s_1 = 0.3$ and  $s_2 = 0.4$,  \textit{moderate shortcut} as $s_1 = 0.1$ and  $s_2 = 0.2$, and \textit{easy shortcut} as $s_1 = 0.4$ and  $s_2 = 0.5$.
\label{filter_descriptions}

\end{document}

%% file: mlhc-submission-files/text/abstract_into_related_work.tex
\begin{abstract}
While deep learning has shown promise in improving the automated diagnosis of disease based on chest X-rays, deep networks may exhibit undesirable behavior related to \textit{shortcuts}. This paper studies the case of spurious class skew in which patients with a particular attribute are spuriously more likely to have the outcome of interest. For instance, clinical protocols might lead to a dataset in which patients with pacemakers are disproportionately likely to have congestive heart failure. This skew can lead to models that take \textit{shortcuts} by heavily relying on the biased attribute. We explore this problem across a number of attributes in the context of diagnosing the cause of acute hypoxemic respiratory failure. Applied to chest X-rays, we show that i) deep nets can accurately identify many patient attributes including sex (AUROC = 0.96) and age (AUROC $\ge$ 0.90), ii) they tend to exploit correlations between such attributes and the outcome label when learning to predict a diagnosis, leading to poor performance when such correlations do not hold in the test population (e.g., everyone in the test set is male), and iii) a simple transfer learning approach is surprisingly effective at preventing the shortcut and promoting good generalization performance. On the task of diagnosing congestive heart failure based on a set of chest X-rays skewed towards older patients (age $\ge$ 63), the proposed approach improves generalization over standard training from 0.66 (95\% CI: 0.54-0.77) to 0.84 (95\% CI: 0.73-0.92) AUROC. While simple, the proposed approach has the potential to improve the performance of models across  populations by encouraging reliance on clinically relevant manifestations of disease, i.e., those that a clinician would use to make a diagnosis.

\end{abstract}

\section{Introduction}

Deep convolutional neural networks (CNNs) have shown promise when applied to chest X-ray classification tasks \citep{Yan2013, giger2018machine}. However, such models often fail to generalize when applied to data from new patient populations, clinical protocols, and institutions \citep{zech2018variable}. In contrast, clinicians are able to adapt to differences in patient populations (e.g., a slightly older population) when moving from one institution to another. We hypothesize that CNNs may be exploiting features \textit{different}  than what a clinician would use to make a diagnosis, in particular easily extracted features that are correlated with the output label in the training population (e.g., the presence of an ECG lead). These features may be useful, even in the test population from the same institution, but if the correlation does not hold across other populations, the model may fail to generalize. The presence of bias in training data leads to potential \textit{shortcuts}. We aim to discourage the learning of these shortcuts and instead encourage the learning of clinically relevant manifestations of disease.  

In this paper, we examine the extent to which potential biases related to age, sex, race, body mass index (BMI), different treatments, and image preprocessing can be exploited by CNNs. Surprisingly, we find that sensitive attributes such as age, sex, and BMI can be learned with AUROC $\ge$ 0.90 using a single chest X-ray. In the context of a diagnostic task, when these attributes appear as spurious correlations in the training data, but not in the test data, model performance with respect to the diagnostic task of interest (e.g., diagnosing heart failure) is substantially affected: the model relies on such correlations during training but is unable to exploit those relationships at test time. In the presence of extreme skew (e.g., perfect correlation between an attribute and a diagnostic task), the model often ends up predicting the attribute rather than the task of interest. 

To mitigate the use of such shortcuts during training, we show that a simple transfer learning approach is surprisingly effective at reducing the negative effects of spurious correlations in the training data. The approach first learns all of the parameters of a deep network on an `unbiased' source task (in our examples, pneumonia); this is followed by learning {\it only} the last layer (a linear model with a few thousand parameters) on the target label with biased data (in our examples, congestive heart failure). This mimics how a clinician might reuse common features (e.g., presence or absence of fluid in the lungs) for different diagnoses. In addition to being simple and effective, this approach has the benefit of not requiring knowledge of the true target class skew. While transfer learning is a well-explored area, it is an empirically surprising result that the last layer simultaneously has enough capacity to learn an expressive model, but not enough capacity to overfit to the bias. 

We evaluate the proposed approach on a population of patients with acute hypoxemic respiratory failure (AHRF). Over two million patients are hospitalized with acute cardiopulmonary conditions in the US each year, and those with AHRF are most commonly diagnosed with one or more of the following etiologies \citep{laribi2019epidemiology}: pneumonia, chronic obstructive pulmonary disease (COPD), and congestive heart failure (CHF). However, determining the underlying causes of AHRF can be clinically challenging, as laboratory testing \citep{roberts2015diagnostic, self2017procalcitonin} and chest X-ray \citep{hagaman2009admission,esayag2010diagnostic} results may be non-specific and difficult to interpret. As a result, patients often receive incorrect initial treatment \citep{ray2006acute}. As chest X-ray imaging is essential for determining the diagnosis, there is potential for deep learning models trained on imaging studies to aid clinicians in making this diagnosis. However, due to the diversity of patients that are hospitalized for acute cardiopulmonary conditions, we require models that generalize well across different demographics.

\subsection*{Generalizable Insights about Machine Learning in the Context of Healthcare}

This work investigates the extent to which deep models can exploit shortcuts in chest X-ray datasets and approaches for mitigating this behavior. Approaches that are robust to spurious correlations will ultimately lead to improvements in the generalizability of models across patient populations and institutions. In this work, we find that a relatively simple method of transfer learning can be surprisingly effective in improving the generalization of models trained on skewed datasets. Our contributions can be summarized as follows: 

\begin{itemize}  \setlength{\itemsep}{-5pt}
  \item we show that deep models can predict patient demographics, treatment, and even image preprocessing steps based on a chest X-ray alone with good performance;  
  \item we show that when training data are skewed with respect to demographics and treatment, such shortcuts can be exploited, leading to biased models that generalize poorly; and
  \item we present a simple, yet effective, approach for reducing, at least in part, such biases. 
\end{itemize}

We apply a transfer learning approach in which we first train a deep learning model to learn pneumonia from an unskewed dataset. Then, we use the model features to predict congestive heart failure in a skewed dataset. This was inspired by how clinicians ``transfer" features among diagnostic tasks. For example, a clinician might look for the presence or absence of fluid in the lungs to aid in distinguishing between congestive heart failure and pneumonia. We find this simple approach, in which we apply a linear classifier to the features of a CNN, works surprisingly well (especially since pneumonia can present differently than congestive heart failure on a chest X-ray). These findings are promising, and can motivate researchers to continue using clinician behavior to guide their development of deep learning models.

\section{Background and Related Work}

\subsection{Problem Setup \& Definitions}
\label{notation}

We consider a supervised learning task, in which we aim to infer a patient's diagnosis based on a chest X-ray. We assume access to a labeled training set, and a separate held-out test set. In this work, we focus on a type of \textbf{\textit{bias}} that arises when training data are skewed with respect to a particular \textbf{\textit{attribute}} (i.e., a characteristic that divides patients into subgroups), resulting in undesirable model performance when the model is applied to test data that do not exhibit the same kind of skew. \textbf{\textit{Skew}} arises when there is a difference in class proportions among patients with and without an attribute (e.g., a greater proportion of patients with heart failure are obese in the training data). This presents a potential \textbf{\textit{shortcut}} for the model. Rather than focusing on clinically relevant manifestations of disease (e.g., an enlarged heart for diagnosing heart failure), the model may focus on irrelevant features (e.g., body fat) when estimating the probability of disease. When applied to a new patient population without a similar relationship between the disease and attribute, the model will generalize poorly. Note, however, that not all skew leads to models that do not generalize. If the skew or correlation holds across populations or institutions (e.g., in general older adults are a greater risk of pneumonia) then learning these features may be informative and contribute to overall model performance. In this work, we focus on spurious correlations that do not necessarily hold across populations. 

\subsection{Related Work}

Recent advancements in deep learning applied to chest X-rays are due in part to the availability of large annotated chest X-ray datasets \citep{Yan2013, giger2018machine, irvin2019chexpert, johnson2019mimic, wang2017chestx}. However, the availability of data alone does not ensure good generalization performance. Zech \textit{et al.} trained CNNs on data from two institutions and showed that they failed to generalize well to a third institution \citep{zech2018variable}. We hypothesize that this poor generalization performance is due in part to CNNs exploiting features that may not be directly related to the underlying diagnoses (i.e., features a clinician would not use for diagnosis), but are only spuriously correlated (e.g., the sex of the patient). This is just one example of the many types of biases that can arise in machine learning. In healthcare, the concept of bias has been explored in different contexts \citep{Gianfrancesco2018, Rajkomar2018, Gijsberts2015, Soboczenski2019}. Most recently, Obermeyer \textit{et al.} found evidence of racial bias in an algorithm widely used in the US healthcare system, reporting that Black patients were assigned the same level of risk as White patients, but were sicker \citep{Obermeyer447}. In contrast, here we focus on a type of bias that arises when there are spurious correlations in the training data that do not necessarily hold in the test set. In this setting, a CNN may exploit these correlations during training, but may fail when applied to the test data.



This biased behavior, which we refer to as the CNN taking a `shortcut' in its prediction, is not confined to the healthcare setting. Much work has been done on understanding when CNNs are biased and can take shortcuts. For example, Wei \textit{et al.} trained a classifier to determine whether a video sequence is playing forward or backward, but realized that the classifier could learn artificial cues (i.e., cues introduced during video production) rather than real world cues \citep{wei2018learning}. Similarly, Doersch \textit{et al.} found that CNNs could latch onto chromatic aberrations in still images rather than learning the desired high-level semantics of images \citep{doersch2015unsupervised}. On the task of discovering representation flaws caused by potential dataset bias, Zhang \textit{et al.} mined latent relationships between representations of image attributes learned by a CNN and compared them to ground-truth attribute relationships, allowing them to spot when a CNN failed in terms of determining non-meaningful relationships between attributes \citep{zhang2018examining}. We consider potential shortcuts in a diagnostic task based on a chest X-ray related to demographics, treatment, and image processing, and explore approaches for preventing a CNN from exploiting those shortcuts.

To improve generalization performance, one can resample the training data to more closely mimic the test data.  For example, Zech \textit{et al.} resampled their training data to ensure equivalent pneumonia prevalence across institutions. However, this approach is unlikely to scale across multiple institutions, especially in settings where the diagnostic task prevalence is unknown \textit{a priori}, or when there is not enough data to allow for resampling. Additionally, one may only have access to a trained model, and not the underlying training data. Thus, we explore an alternative approach to resampling based on transfer learning \citep{pan2009survey}. Transfer learning, and more specifically pretraining, in which one initializes the weights of a model for the \textit{target task} using data from a related \textit{source task}, has had wide success across many domains \citep{imagenet_cvpr09, sharif2014cnn}. In the context of medical imaging, Shin \textit{et al.} successfully transferred features from a model trained on ImageNet to the task of classifying the presence of interstitial lung disease \citep{Shin2016}. However, if the tasks are not related, transferring from source to target can result in \textit{negative} transfer, leading to worse performance. For example, Raghu \textit{et al.} showed that on the task of diagnosing diabetic retinopathy based on images of the back of the eye, pretraining on ImageNet did not improve performance relative to random initialization \citep{raghu2019transfusion}. In this work, we consider a source task that is, in part, related to the target task but does not exhibit the same spurious correlations.


Related to transfer learning, multitask learning shares knowledge among tasks by learning tasks \textit{jointly} \citep{caruana1997multitask}. Multitask learning has shown promise across a vast array of applications including healthcare \citep{aguilar2019modeling, mccann2018natural, verma2020using, wang2020environment, radwan2018vlocnet++}. For example, Kyono \textit{et al.} trained a CNN on mammograms to jointly predict both a cancer diagnosis as well as patient features from radiological assessments known to be associated with cancer \citep{kyono2019multi}. Similarly, Suresh \textit{et al.} showed that learning patient subgroups and then predicting outcomes for separate patient populations in a multitask framework led to better predictive performance of in-hospital mortality across groups and overall \citep{suresh2018learning}. In our application, multitask learning could potentially help mitigate bias by sharing information across the source and target tasks. However, nothing is preventing the model from \textit{not} sharing information and exploiting shortcuts in either task. Additionally, it assumes access to all labeled training data, whereas a transfer learning approach can leverage stored parameters. Given the difficulty of sharing data across institutions, such an approach may not be feasible. Nonetheless, for completeness we explore a multitask setup as an alternative to the transfer learning approach.

%% file: mlhc-submission-files/text/proposed_approach.tex
\section{Preventing Shortcuts Using Transfer Learning}

In this section, we formalize the problem of learning shortcuts from a chest X-ray. We then present an approach that aims to mitigate the learning of these shortcuts when the target task exhibits skew in the training data, i.e., the diagnostic outcome is highly correlated with some clinically irrelevant attribute. This approach is based on the idea of transfer learning, and re-uses the features learned in a source task to learn an updated model for the target task.

\subsection{Formalization and Notation}  Given a chest X-ray image  $X\in \mathbb{R}^{d\times d}$, we consider the task of predicting some {\it target} diagnosis $y_t \in \{0,1\}$. We assume a convolutional neural network (CNN) $f(e(X;\thetaB);\wB)$ is the composition of two functions: an encoder $e$ parameterized by $\thetaB$ and consisting of interleaved convolutions and nonlinearities, as well as a linear layer parameterized by $\wB$ and consisting of an affine transformation followed by a sigmoid activation. The CNN takes as input the chest X-ray image $X$ and outputs the probability of the outcome of interest for the given chest X-ray, $\hat{y_t} \in [0,1]$. Note that vast majority of the parameters of the network are usually in the encoder (i.e., $\textrm{dim}(\wB) \ll \textrm{dim}(\thetaB)$).

Given a training set composed of $n$ labeled chest X-rays $\{X^{(i)}, y^{(i)}_t\}_{i=1}^n$, we aim to learn the parameters (i.e, $\thetaB, \wB$) of a CNN that maps $X$ to $y_t$ by minimizing a loss function $\mathcal{L}$ that quantifies how well a prediction matches the ground truth (e.g., binary cross-entropy loss). In addition to the input image and target output label, for each individual $i$, we assume access to an unbiased {\it source} dataset $\{y^{(i)}_s\}_{i=1}^n$ that we use during training, or a model with its parameters \textit{pretrained} on an unbiased source dataset. In our analysis, we assume the spurious correlation is modeled by a bias attribute $b \in \{0,1\}$ (e.g., age$>$63), which may be spuriously correlated with the outcome $y_t$, and is not typically considered clinically relevant in diagnosing $y_t$.

We consider a scenario in which in the training and validation data associated with the target task $y_t$ and attribute $b$ may be highly correlated, but this correlation does not hold in the test data. This represents a setting where there is some skew in the training and validation sets that is not present in the test set. Using a standard approach of training a model to predict $y_t$, the model could simply learn to predict $b$. However, we want to avoid learning $b$, because it is of little clinical relevance to the task and as a result, there may be no correlation between $y_t$ and $b$ in the test set. We aim to develop an approach that encourages the model to learn features that predict $y_t$, not $b$.

\subsection{Training Scheme \& Architecture}\label{architecture}

We hypothesize that the chest X-ray features used to predict $y_s$, the `unbiased' or curated source task, can also be used to predict $y_t$.  Thus, we consider a simple approach, in which we first learn to predict $y_s$. We then freeze the associated encoder parameters $\thetaB$, and tune the output layer, $\mathbf{w}$, to predict the target task label $y_t$. Since $y_s$ is not correlated with attribute $b$, we anticipate that the model will not learn features associated with $b$; thus eliminating the possibility of a shortcut when used in predicting $y_t$.

\begin{figure}[htb!]
    \centering
    \includegraphics[width=\linewidth]{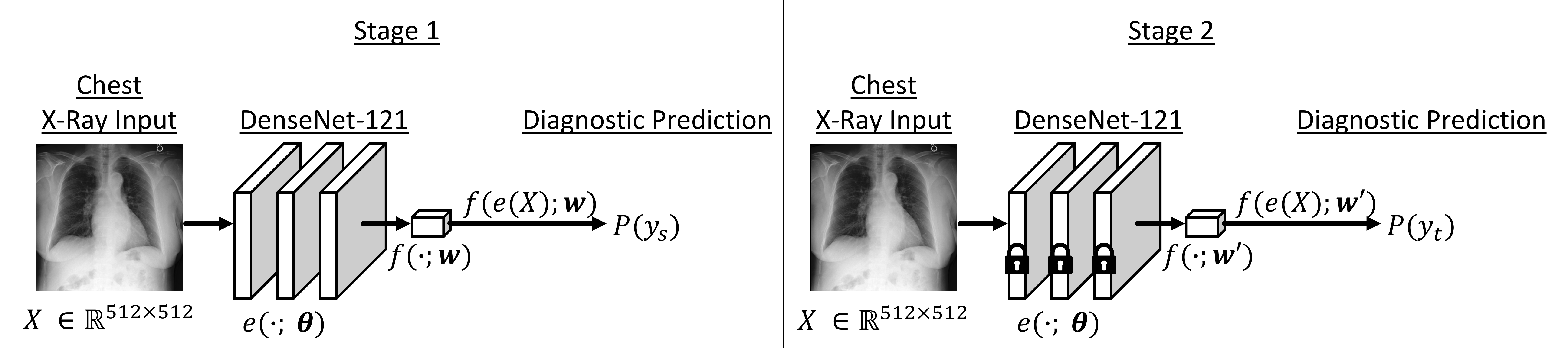}
    \caption{Training scheme. The CNN consists of an encoder $e$, a DenseNet-121, followed by a final linear layer $f$ parameterized by $\mathbf{w}$. First, we train the model to predict $y_s$, the curated source task. Then, we freeze the encoder parameters $\thetaB$ and tune weights $\mathbf{w}$ to predict $y_t$, the target task.}
    \label{fig:architecture}
\end{figure}

Our training scheme (shown in \textbf{Figure \ref{fig:architecture}}) is the standard transfer-learning approach and has two stages: full training on the source task and last-layer fine tuning on the target task. In stage 1, we pretrain a model, including all layers, to predict the source task, or solving 
$\thetaB^*, \wB_s^* = \arg \min_{\thetaB,\wB} \sum_{i} \mathcal{L}(f(e(X^{(i)};\thetaB);\wB),y_s^{(i)})$ via a standard descent method. The optimal encoder parameters $\thetaB^*$ are retained and the parameters $\wB_s^*$ for the last layer for the source task are discarded. In the second stage, we then learn the parameters of the last layer to predict the target task while holding the parameters for the encoder constant at $\thetaB^*$, solving
$\wB_t^* = \arg \min_{\wB} \sum_i \mathcal{L}(f(e(X^{i};\thetaB^*);\wB),y_t^{(i)})$ by a standard descent method. The resulting classifier is  $f(e(\cdot;\thetaB^*);\wB_t^*)$, combining the encoder parameters learned on the source task and the linear layer learned on the target task.

For the CNN, we use a common CNN architecture, DenseNet-121, with a final linear layer $f$ of size 1024. We chose DenseNet-121 because it has been shown to outperform other CNN architectures on tasks involving chest X-rays \citep{irvin2019chexpert}. DenseNet-121 connects each layer to every other layer in the network, as opposed to connecting each layer only to its subsequent layer. These connections allow for stronger feature propagation and feature reuse, which reduce the number of parameters needed in the network.

%% file: mlhc-submission-files/text/experiments.tex
\section{Experimental Setup}

To test the ability of a simple transfer learning approach to mitigate bias, we consider a setting in which we inject bias into a diagnostic task (i.e., we introduce potential shortcuts). We introduce this bias through (i) resampling the data such that potential sensitive attributes (like race) are highly correlated with the outcome, and (ii) additional image preprocessing steps based on the label. On held-out test data, we compare the model's discriminative performance to the performance of a model trained directly on the biased target task training data. In addition, we include comparisons to transfer learning with data from MIMIC-CXR and CheXpert compared to data from our carefully curated source task. Finally, we perform a comparison to a multitask approach and a sensitivity analysis.

\subsection{Datasets} 
\label{dataset}
\par \noindent \textbf{AHRF dataset.} We considered a cohort of 1,296 patients admitted to a large acute care hospital center during 2017-2018 who developed AHRF. We analyzed chest X-rays acquired closest to the time of AHRF onset. Each patient had a corresponding study, containing one or more chest X-rays taken at the same time. Within each study, we considered only frontal images as shown in \textbf{Figure \ref{fig:frontal_x_ray}}. This study was approved by the Institutional Review Board at the University of Michigan. 

Chest X-ray images were obtained from a picture archiving and communications system in the form of digital imaging and communications in medicine files and preprocessed as follows. Solid black borders were removed from the original images. Histogram equalization was applied to the images to increase contrast in the images. Images were originally as large as $2000 \times 3000$ pixels, but were resized such that their smaller axis was set to $512$ pixels while preserving their aspect ratio. This allowed for subsequent cropping (horizontally or vertically) to produce a $512 \times 512$ image.

We consider the tasks of learning to diagnose patients with congestive heart failure (CHF) (as our target task) and pneumonia (as our source task). All patients were labeled by at least two clinicians after reviewing the hospital chart and X-ray images. Clinicians assessed each patient with a scale of 1-4 for each diagnosis, with 1 being very likely and 4 being very unlikely. Scores were averaged across clinicians and thresholded at 2.5.

For each patient, we extracted several attributes that could serve as $b$, including body mass index (BMI), age, sex, race, and the presence of a pacemaker. Demographic variables were extracted from the electronic health record, while images were manually labeled for the presence of a pacemaker.  We dichotomized each variable:  i) BMI $\ge$ 30, ii) age $\ge$ 63 years (the median age of our dataset), iii) sex $=$ female, iv) race $=$ Black, and v) pacemaker present. In addition to these demographic and treatment variables, in a series of follow-up experiments (\textbf{Section \ref{preprocessing}}), we inject a `synthetic' bias via an image preprocessing step. More specifically, we apply a Gaussian filter to each chest X-ray. Such filters are commonly used to reduce noise. Here, instead of applying an identical filter to all images, we mimic possible heterogeneity in preprocessing pipelines by varying the standard deviation of the filters. 

\par \noindent \textbf{MIMIC-CXR and CheXpert datasets.} We also considered the MIMIC-CXR and CheXpert datasets \citep{johnson2019mimic,Goldberger2000, https://doi.org/10.13026/c2jt1q, irvin2019chexpert}, two large publicly available chest X-ray datasets. Similar to the AHRF cohort, only frontal images were considered. The images are provided with 14 labels text-mined from radiology reports, categorized as positive, negative, uncertain, and not mentioned. Following Irvin and Rajpurkar \textit{et al.}, we mapped uncertain labels to positive. 

The images are also provided with several attributes that could serve as $b$. Specifically, CheXpert was provided with age and sex, which we dichotomized the same way as described above. MIMIC-CXR was provided with age, sex, insurance provider, and marital status. We dichotomized each variable: i) age $\ge$ 60 years (since we were provided deciles, e.g.,``50-60"), iii) sex $=$ female, iv) insurance $=$ Medicaid, and v) marital status $=$ married.




\subsection{Baselines and Proposed Approach} \label{baselines}
We compare the transfer learning approach, \texttt{Last Layer M/C + A}, to other approaches (\textbf{Table \ref{approaches}}) with the goal of answering the following questions: 

\begin{itemize}
    \item Does directly training on the target diagnostic task exploit shortcuts? 
    \item Does pretraining on data from MIMIC-CXR and CheXpert, 2 large chest X-ray databases, prevent the exploitation of shortcuts? 
    \item Does pretraining on a carefully curated source task, in addition to MIMIC-CXR and CheXpert, further prevent the exploitation of shortcuts?
\end{itemize}

Before any training (and pretraining), we initialized all models on ImageNet. We refer to each approach based on the pretraining and fine-tuning approach: 
\begin{table}[!ht]
\centering
\caption{We refer to each approach based on the pretraining and fine-tuning approach. \texttt{All Layers} fine tunes all layers of an ImageNet initialized model. Second, \texttt{All Layers M/C} pretrains on MIMIC-CXR and CheXpert, and then fine tunes all layers on the biased target task. \texttt{Last Layer M/C} also pretrains on MIMIC-CXR and CheXpert, but fine tunes only the last layer. Finally, \texttt{Last Layer M/C + A} pretrains on MIMIC-CXR and CheXpert, the network trains to predict an unbiased, curated source task, and then tunes the last layer on the biased target task.}
\begin{tabular}{c|>{\centering\arraybackslash}p{5cm} c}

\toprule
\label{approaches}

\centering Approach &  \multicolumn{2}{c}{Description} \\
\midrule
    & \centering \underline{Pretraining} & \underline{Fine-tuning}\\
 \textbf{\texttt{All Layers}}     & none* & all layers \\
 \textbf{\texttt{All Layers M/C}}     & MIMIC-CXR and CheXpert  & all layers \\
 \textbf{\texttt{Last Layer M/C}}      & MIMIC-CXR and CheXpert &  last layer \\
 \textbf{\texttt{Last Layer M/C + A}}     & MIMIC-CXR and CheXpert followed by AHRF unbiased source task &  last layer\\
\bottomrule
\end{tabular}

\begin{tablenotes}

      \small
      \item \hspace{1cm} *note: as is common, all approaches pretrain on ImageNet.
    \end{tablenotes}
\end{table}
\\
\par \noindent \textbf{All Layers.} We first consider an approach in which a model, initialized on ImageNet, is trained by minimizing the cross-entropy loss associated with the target task labels $y_t$. Though we initialize on ImageNet, we refer to this approach as `no pretraining' since it does not use any domain-specific data (i.e., chest X-rays) for pretraining. Given the correlation between $b$ and $y_t$, we expect that the network will learn to exploit shortcuts related to $b$. This in turn will lead to poor generalization performance on a test set in which $y_t$ is not correlated with $b$. 

\par \noindent \textbf{All Layers M/C.} Second, we consider an approach in which we initialize on MIMIC-CXR and CheXpert by pretraining on all 14 provided labels (instead of ImageNet).  Initializing on pretrained weights from large chest X-ray databases could encourage the model to learn more clinically relevant features. However, we measured the correlation between all 14 tasks and attributes $b$ (\textbf{Appendix Table \ref{tab:correlations}}) and found many of the tasks to be correlated with attributes $b$. Moreover, as above, all layers are tuned on the biased training data. Thus, there is still a significant possibility that the model will learn to take the shortcut. 

\par \noindent \textbf{Last Layer M/C.} Finally, we consider a transfer learning approach that relies entirely on the MIMIC-CXR and CheXpert datasets. Here, we initialize on MIMIC-CXR and CheXpert by pretraining on all 14 provided labels and then, as in the proposed approach \texttt{Last Layer M/C + A}, freeze all but the last layer, tuning the output layer, $\mathbf{w}$ by minimizing the cross-entropy loss on the biased target task data. This is similar to the proposed approach, but does not make use of an additional \textit{curated} source task. We hypothesize that such an approach will, in part, mitigate the use of shortcut features. However, in addition to the correlations between the 14 MIMIC-CXR and CheXpert tasks and attributes $b$ (\textbf{Appendix Table \ref{tab:correlations}}), since patients with AHRF are generally sicker than the patients found in MIMIC-CXR and CheXpert, we hypothesize that simply pretraining on these two datasets will not perform as well as the transfer learning approach using the curated source task dataset. 

\subsection{Training Details and Evaluation} 

All of our code is publicly available to allow for replication\footnote{https://gitlab.eecs.umich.edu/mld3/deep-learning-applied-to-chest-x-rays-exploiting-and-preventing-shortcuts}. We partitioned the data into train, validation, and held-out test sets based on patients. Splitting by patient is important since each patient could have more than one chest X-ray in a single study (on average, each patient had 1.07, 3.84, and 2.95 images for AHRF, MIMIC-CXR, and CheXpert, respectively).  To generate a single prediction for each patient, we averaged the predictions over all images in a study. 

\par \noindent \textbf{AHRF dataset training setup.} We trained each model using stochastic gradient descent with momentum and a batch size of 32. We minimized the binary cross-entropy loss. Hyperparameters, including the learning rate ($\in$ \{$10^{-3}$, $10^{-2}$, $10^{-1}$\}) and momentum ($\in$ $\{0.8,0.9\}$), were selected based on the validation data, optimizing for the area under the receiver operating characteristics curve (AUROC). To augment the training data, images in the training set were randomly cropped to $512$ on either side and augmented with random in-plane rotations up to $15$ degrees. Validation and test images were center cropped to $512 \times 512$. We initialized models on either ImageNet or by pretraining on the MIMIC-CXR and CheXpert datasets. 

\par \noindent \textbf{MIMIC-CXR and CheXpert datasets training setup.} We applied the same preprocessing to these data as described above. We optimized the sum of the \textit{masked} binary cross-entropy loss across labels, masking the loss for labels with a missing value. Following Irvin \textit{et al.}, we used Adam with default $\beta$-parameters of $\beta_1 = 0.9, \beta_2 = 0.999$, learning rate $1 \times 10^{-4}$, and a batch size of 16 \citep{irvin2019chexpert}. We trained for 3 epochs with 3 different random initializations, saving checkpoints every 4,800 batches. When pretraining for the AHRF cohort, we combined these datasets and then selected the checkpoint that performed the best on the CheXpert validation set, measured by average AUROC across all 14 labels. We used the same training procedure when training models on the datasets separately.

\par \noindent \textbf{Evaluation.} Our primary quantitative measure of performance on held-out test data is the AUROC, denoted by AUROC($\hat{y}, y_t$). We computed empirical 95\% confidence intervals using 1,000 bootstrapped samples of the held-out test set. For some experiments, we also computed this metric considering attribute $b$ as the ground truth to test to what extent the model is simply predicting $b$, and denote this as AUROC($\hat{y}, b$). To compare models, we calculated statistical significance using a resampling approach, subtracting the AUROC scores from 1,000 pairs of bootstrapped sampled of the held-out test set and report the percent of times it falls below 0. We also visualize the part of the chest X-ray image that the model uses to make a prediction using Gradient-weighted Class Activation Maps (GradCAM) \citep{selvaraju2017grad}. This involves taking a weighted sum of the gradient of the final linear layer of the network and the 1024 feature maps, and then upscaling this result to the size of the chest X-ray.

\section{Experiments and Results}

In the experiments that follow we seek to answer the following questions:

\begin{itemize}
    \item  How easy is it to infer demographics (e.g., age), treatment (e.g., pacemaker), and subtle differences in preprocessing steps (e.g., filter parameters) based on chest X-rays? (\textbf{Sections \ref{infer bias} \& \ref{infer_Gaussian}})
    \item  If the training data have spurious correlations, will a CNN exploit features related to these correlations in lieu of relying on the actual clinical presentation of disease when making diagnostic predictions? (\textbf{Sections \ref{shortcuts exploited} \& \ref{infer_Gaussian}}) 
    \item  How effective is the presented transfer learning approach in mitigating this type of bias and how does it compare to alternative approaches? (\textbf{Section \ref{2-Stage} \& \ref{infer_Gaussian}})
    \item How does the effectiveness of the proposed approach change when the source task is no longer ``unbiased''? (\textbf{Section \ref{infer_Gaussian}})
\end{itemize}

We first answer a subset these questions in the context of demographic and treatment biases, and then in a series of follow-up experiments repeat our analyses with a synthetic bias injected via an image preprocessing step, answering the final question.

\subsection{Experiments with Demographics/Treatment Shortcuts}

First, we aim to understand how easy it is to infer demographics and treatment based on chest X-rays across datasets and the tendency of a model to exploit these features when correlated with the outcome of interest. We then test the ability of the transfer learning approach in mitigating the use of these features, and compare to multiple baselines. 

\subsubsection{Inferring Demographics and Treatment Based on Chest X-rays} 
\label{infer bias}
To establish that shortcuts can be easily exploited by models, we first measure the extent to which models can learn potentially sensitive or clinically irrelevant attributes (e.g., race). Here, instead of training a model for the diagnostic tasks, we aim to learn a mapping from the chest X-ray image to each attribute $b$. For the AHRF cohort, we randomly sampled 20\% of the data in which no patients were missing BMI, age, sex, race, or pacemaker information as the test set. We trained on 80\% of the remaining patients and validated on 20\%. We initialized these models on MIMIC-CXR and CheXpert. For the MIMIC-CXR and CheXpert cohorts, we randomly sampled 1000 patients for both the validation and test sets, and trained on the remaining patients. We sampled the test/validation patients such that no patients were missing the relevant attributes. These models were initialized on ImageNet.

\begin{table}[!ht]
\centering 
\caption{CNNs perform surprisingly well at inferring attributes like age and sex from chest X-rays. The models achieve near-perfect performance on identifying sex ($=$F) across all datasets and identifying age ($\ge 63$ for AHRF, CheXpert, $\ge60$ for MIMIC-CXR) on two out of the three datasets. The models achieve similarly good performance on BMI and pacemaker in AHRF. The models do worse, but above chance, on binary attributes for race ($=$Black), insurance ($=$Medicaid), and marital status ($=$Married), although we suspect these are explained by other correlations in the data. We aim to prevent a model from using these attributes as shortcuts.}
\label{tab:shortcut_splits_results}
\scalebox{0.75}{
\begin{tabular}{ccccccc}

\toprule &   \multicolumn{2}{c}{ \textbf{AHRF}} &   \multicolumn{2}{c}{ \textbf{MIMIC-CXR}} &   \multicolumn{2}{c}{ \textbf{CheXpert}}\\ 

\textbf{Task} &   \textbf{\% pos} & \textbf{AUROC (95\% CI)} &   \textbf{\% pos} & \textbf{AUROC (95\% CI)} &   \textbf{\% pos} & \textbf{AUROC (95\% CI)} \\ 
\midrule 
\textbf{Age}  &  55 &  0.72 (0.66-0.78) &  57 &  0.90 (0.89-0.91) &    56 &  0.95 (0.94-0.95)  \\ 
\textbf{Sex}  &  40 &  0.96 (0.94-0.98) &   46 &  1.00 (1.00-1.00) &    41 &  1.00 (1.00-1.00)  \\ 
\textbf{BMI}  &  44 &  0.91 (0.88-0.94) &   -- &  -- & -- &  -- \\ 
\textbf{Race}  & 9 & 0.66 (0.54-0.79) &   -- &  -- & -- &  -- \\ 
\textbf{Pacemaker}  &  9 &  0.97 (0.91-1.00) &  -- &   -- &  -- &  -- \\ 
\textbf{Insurance} &  -- &   -- &  9 &  0.70 (0.67-0.72) &  -- &   -- \\
\textbf{Marital}  &   -- &   -- &  44 & 0.65 (0.63-0.66) &  --  &  -- \\ 
\bottomrule 
 \hspace{\fill}
\end{tabular}
}
\end{table}

\par
\begin{figure}[!ht]
    \centering
    \scalebox{1}{

    \begin{subfigure}{}
        \includegraphics[width=6cm]{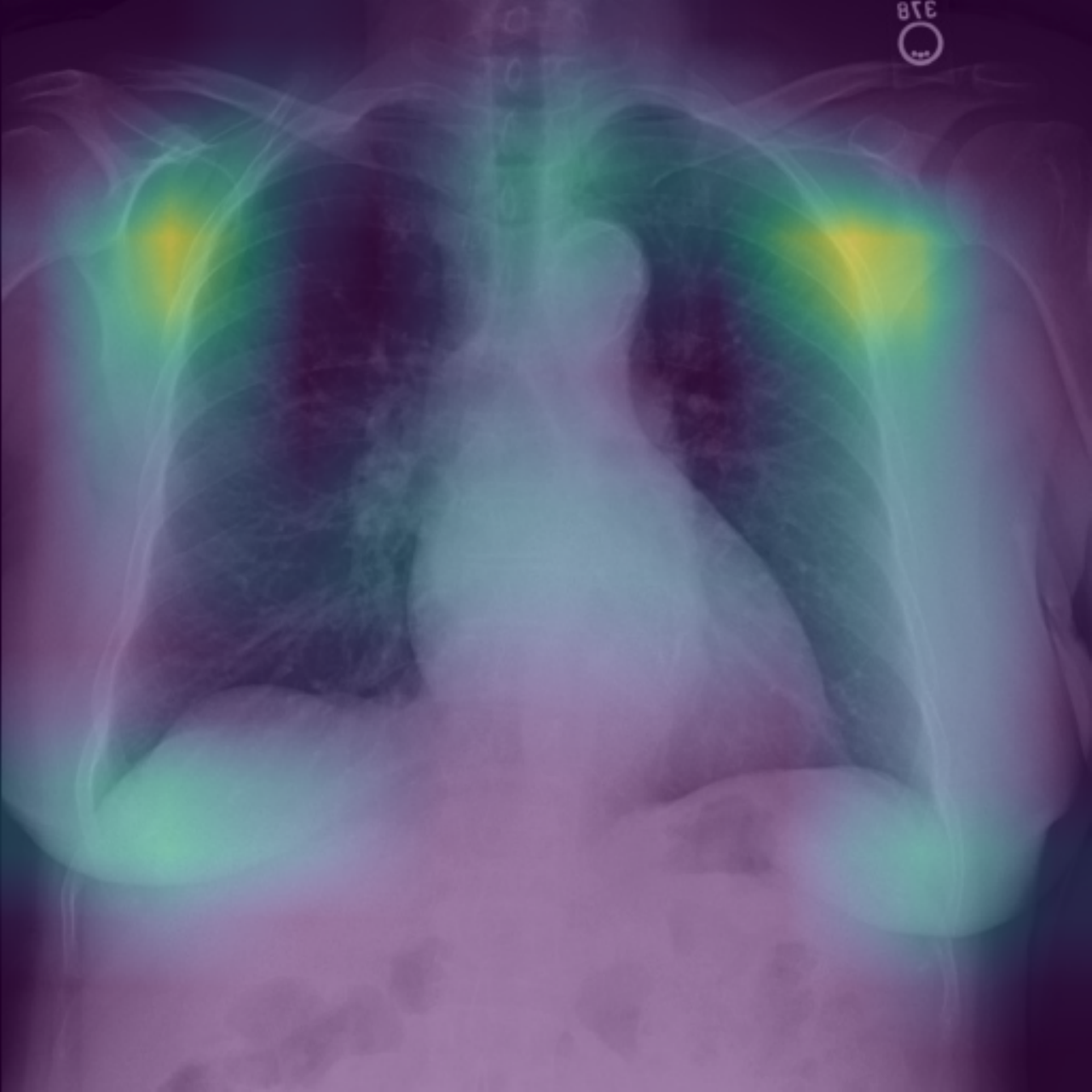}
    \end{subfigure}    
    \begin{subfigure}{}
        \includegraphics[width=6cm]{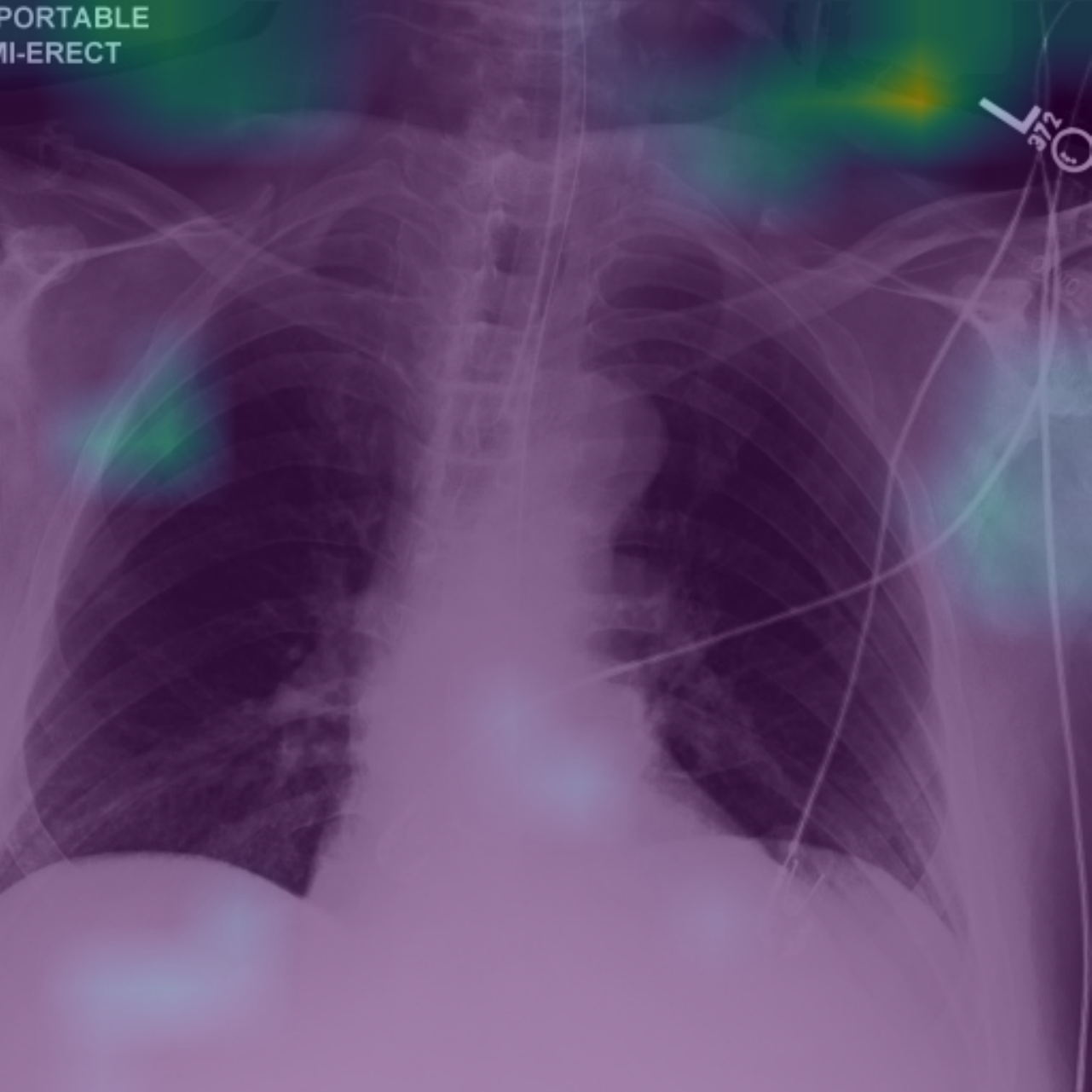}
    \end{subfigure}
    }

    \caption{When trained to predict a patient's sex based on their chest X-ray, the model's GradCAM localizations focus more on breast tissue and the scapulae for female patient (a) compared to a male patient (b).}%
    \label{fig:frontal_x_ray}%

\end{figure}

\begin{figure}[!ht]
   \centering
   
       \begin{subfigure}{}
        \includegraphics[width=6cm]{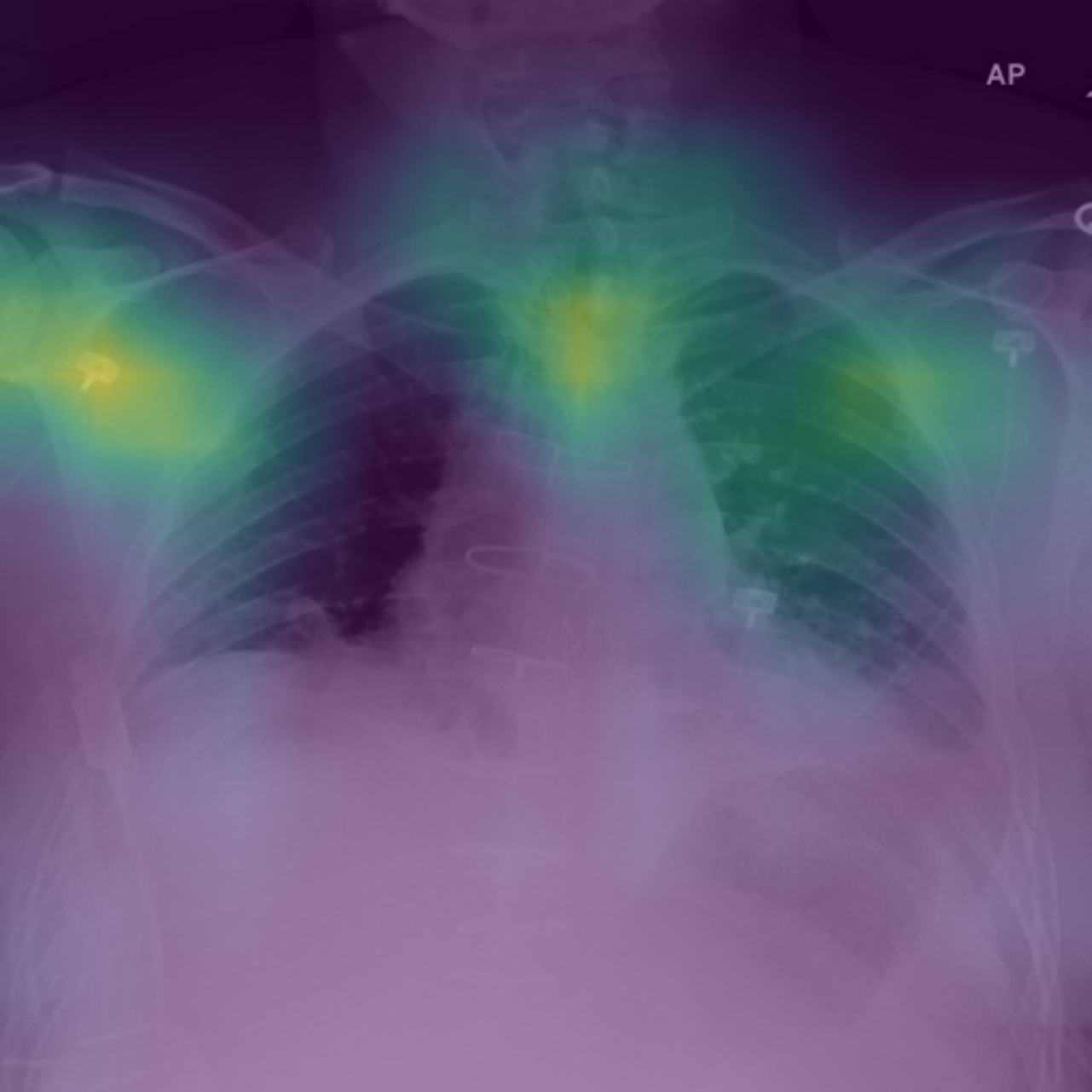}
    \end{subfigure}    
    \begin{subfigure}{}
        \includegraphics[width=6cm]{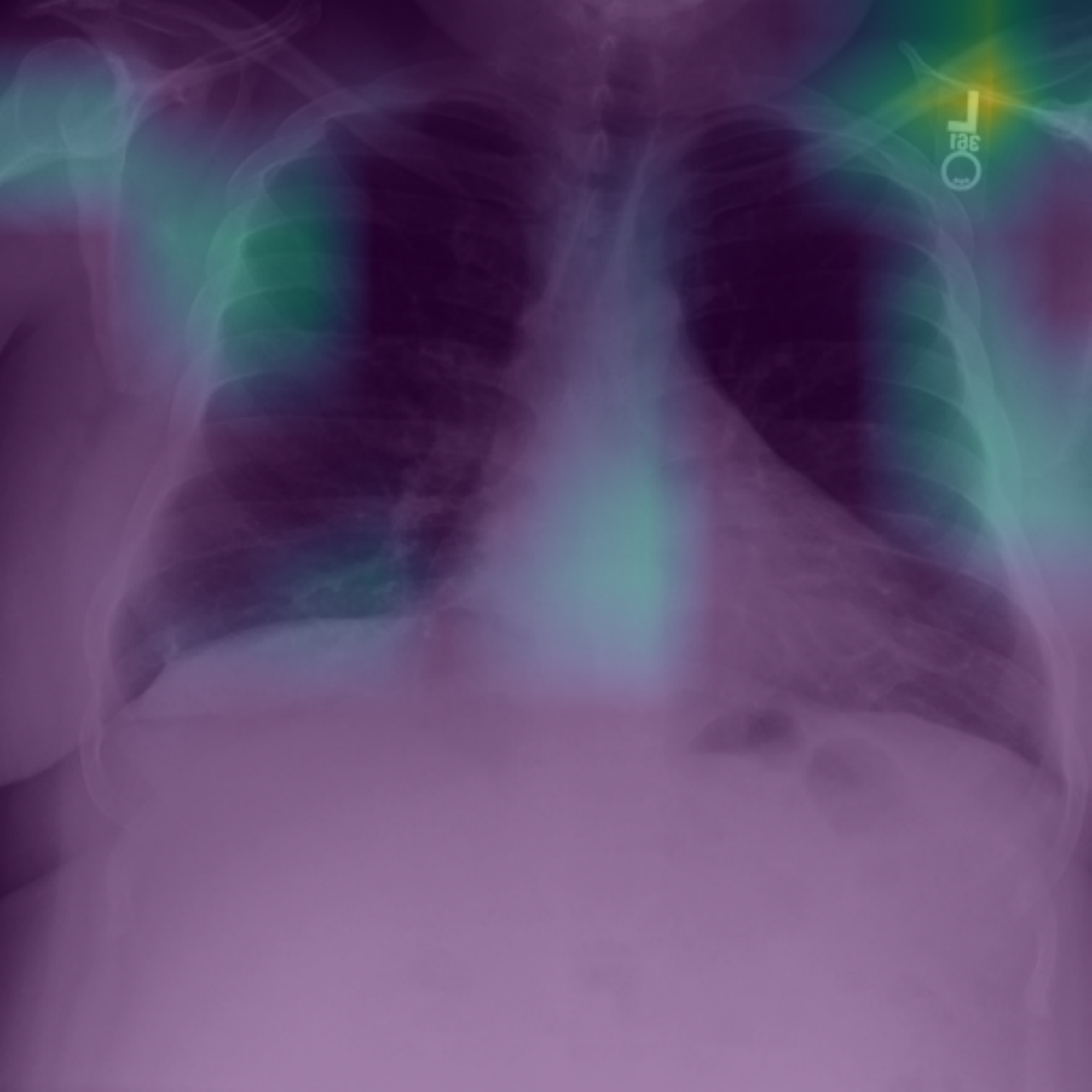}
    \end{subfigure}
    
   \caption{(a) Given a chest X-ray of patient with age$\ge$60, the model localizes around joints and bones, suggesting that it is using bone density in its prediction. (b) When tasked to predict insurance, the model focuses around bone edges and joints, further suggesting that it may be using age as an indicator of insurance status.}
   \label{fig:gradcam}
\end{figure}

 \noindent \textbf{Results and Discussion.} Across all attributes, except for race and marital status, we were able to train a model with good discriminative performance (\textbf{Table \ref{tab:shortcut_splits_results}}). From a chest X-ray alone, we can classify patients as female or not with nearly perfect AUROC across all three datasets. From a clinician's perspective, this is somewhat surprising, since it is difficult for humans. We hypothesized that the model is picking up on the presence of breast tissue, though the GradCAM activations show that the model looks at both breast tissue and the scapulae. (\textbf{Figure \ref{fig:frontal_x_ray}a} vs \textbf{Figure \ref{fig:frontal_x_ray}b}). Additionally, we can learn age with AUROC $\ge$ 0.90 across two of the three datasets. We hypothesize that the model is picking up on bone density (\textbf{Figure \ref{fig:gradcam}a}). In classifying patients as BMI$\ge$30, we also achieved an AUROC $\ge$ 0.90; this is likely due to the visible presence of body fat on a chest X-ray. Unsurprisingly, we can also classify patients based on the presence of a pacemaker from a chest X-ray.

For the MIMIC-CXR cohort, performance was worse in predicting insurance (AUROC=0.70 [95\% CI: 0.67-0.72]) and marital status (AUROC=0.65 [95\% CI: 0.63-0.66]), although there was a signal. We hypothesize that the model is able to extract \textit{some} signal from the images due to the presence of correlation with age (e.g., older patients are more likely to be married or on Medicare). To test this, we calculated the AUROC of predicting each of these attributes using age as the prediction value. Unsurprisingly, we achieved an AUROC of 0.58 (0.58-0.59) and 0.70 (0.69-0.70) for marital status and insurance, respectively. Given this correlation, we hypothesize that the model is able to use features related to age in making these predictions. Furthermore, the GradCAM visualization for insurance in \textbf{Figure \ref{fig:gradcam}b} indicates that the model may be relying on features related to bone density to make a prediction.

On the task of race in the AHRF cohort, performance was significantly worse (AUROC=0.66 [95\%CI 0.54-0.79]), though there is still a signal (i.e., the performance is better than random chance). Again, we hypothesize that this may be due to the presence of a small amount of correlation between race and other attributes in our data. Similar to above, we tested the ability of using other attributes to predict race. We achieved an AUROC of 0.61 (0.56-0.66) using age, 0.55 (0.51-0.60) using sex, 0.52 (0.51-0.60) using BMI, and 0.47 (0.46-0.49) using pacemaker. Thus, there is at least some correlation between age and race. 

Overall, these results (for the most part) show that attributes such as demographics and treatment are easy to infer based on a chest X-ray. Thus, in a setting where a diagnostic task of interest is correlated with one of these attributes, a model could exploit such features despite a lack of clinical relevance. This is especially critical in a setting where a model exploits shortcut features that are correlated with a sensitive attribute (such as insurance, which can be used as a proxy for socioeconomic status). 

\subsubsection{Shortcuts are Easily Exploited} 
\label{shortcuts exploited}

Now that we have established that, even with small amounts of training data  (i.e., $n<1,000$), it is possible to learn a model that can classify individuals according to demographics and treatment based on their chest X-rays, we consider the ability of a model to exploit these features when correlated with an outcome of interest (i.e., the diagnostic task). In other words, to what extent will a model learn to rely on these spuriously correlated features over clinically relevant features. 

Here, we aim to learn a mapping from the chest X-ray image to our target diagnostic task $y_t$, CHF. For illustrative purposes, we consider the extreme case where, in each of the training/validation sets, we introduce a 1:1 correlation between the target task label $y_t$ and some latent attribute $b$ (e.g., age), but no correlation in the test set. To ensure that the models were comparable, we kept the training set sizes and the prevalence of CHF similar (\textbf{Appendix Table \ref{tab:skewed_unskewed_datasets}}). To keep the test set consistent across all tasks, we randomly resampled a test set ($n=111$) in which CHF was not correlated with any of the potential shortcut attributes. 

In addition to the AHRF cohort, we also ran this experiment using the MIMIC-CXR and CheXpert datasets. Since MIMIC-CXR and CheXpert do not have a CHF label, we used pneumonia as the target diagnostic task $y_t$. Despite the change from CHF to pneumonia, we hypothesized that a model will \textit{still} take shortcuts. To keep the test set consistent across all tasks, we randomly resampled a test set ($n = 1000$) in which pneumonia was not correlated with any of the potential shortcut attributes.

We hypothesize that in training the model to directly predict CHF or pneumonia on the skewed data, it will exploit the shortcut features associated with the bias $b$ and generalize poorly when applied to the test set. To measure the extent to which the injected bias affects performance, we compare the learned model's performance, {\tt Skewed Training Data}, to the performance of a model trained on data that more closely mimics the test set, {\tt Unskewed Training Data}. To generate the unskewed training dataset, we subsampled the data so that $b$ was uncorrelated with $y_t$. Again, to facilitate comparisons, we kept the size of the training data similar.

\begin{figure}[!htbp]
    \centering
       \includegraphics[width=\textwidth]{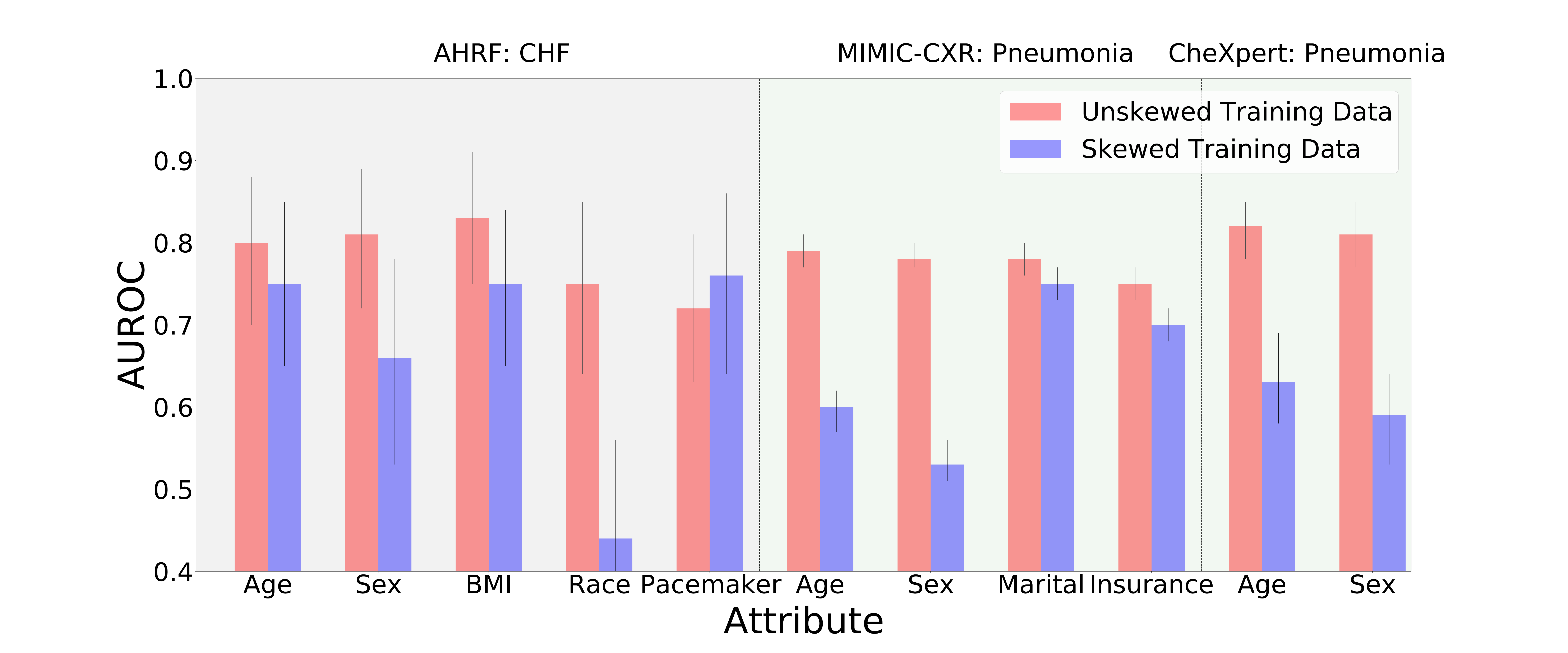}
        
    \caption{We compare, on a dataset with no attribute skew, the performance of {\tt Unskewed Training Data}, a model trained on a dataset without attribute skew; and {\tt Skewed Training Data}, a model trained on a dataset with attribute skew. We report the AUROC with error bars representing a 95\% CI. In the context of age, sex, BMI, race, marital status, and insurance, the model trained on skewed data has a substantial performance drop compared to the model trained on unskewed data.}
    \label{fig:skewed_vs_unskewed}
\end{figure}
\par \noindent \textbf{Results and Discussion.} Across all attributes except pacemaker, performance of the model trained on skewed data was consistently worse compared to a model trained on unskewed data (\textbf{Figure \ref{fig:skewed_vs_unskewed}}). We report statistical significance in \textbf{Appendix Table \ref{tab:statistical_tests_us}}. This suggests that, for the most part, in the presence of skew, a model will learn to exploit shortcut features associated with attribute $b$. The presence of skew towards sex appears to greatly affect performance across all three datasets (AUROC 0.81 vs 0.66, 0.78 vs 0.53, 0.82 vs 0.63 for AHRF, MIMIC-CXR, and CheXpert, respectively). As presented in \textbf{Section \ref{infer bias}}, sex=female is one of the easiest attributes to learn. Thus, in the presence of skew, the model can more easily take a shortcut to learn sex than it can to learn BMI or age. Additionally, for age in the AHRF cohort and marital status and insurance in the MIMIC-CXR cohort, we find that the presence of skew does not \textit{completely} degrade predictive performance of $y$ (i.e., an AUROC of 0.5), which suggests that the model is still able to learn some useful features of $y$. Notably, the model trained on the dataset skewed towards pacemaker does not perform worse than the model trained on the unskewed dataset (the number of positive cases is small and error bars are large). Nonetheless, the drop in performance due to the presence of skew in the other six attributes presents an opportunity to explore approaches that mitigate the presence of skew. 

\subsubsection{Learning Robust Features Using Transfer Learning} 
\label{2-Stage}
Having established that CNNs will take shortcuts if available, we explore the effectiveness of the presented transfer learning approach in mitigating this behavior, comparing to the alternative approaches presented in \textbf{Section \ref{baselines}}. Here, we aim to learn a mapping from the input chest X-ray $X$ to a prediction, $y_t$, representing whether or not the patient has CHF. We assume access to training data (or at least a trained network) for a second task, $\{X^{(i)},y_s^{(i)}\}_{i=1}^n$, that does not contain the same bias (i.e., $y_s$ is not correlated with $b$), or access to the model trained on the second task and its parameters $\thetaB$. We consider a source task corresponding to diagnosing pneumonia. 

\begin{figure}[!ht]
    \centering
    \includegraphics[width=1\textwidth]{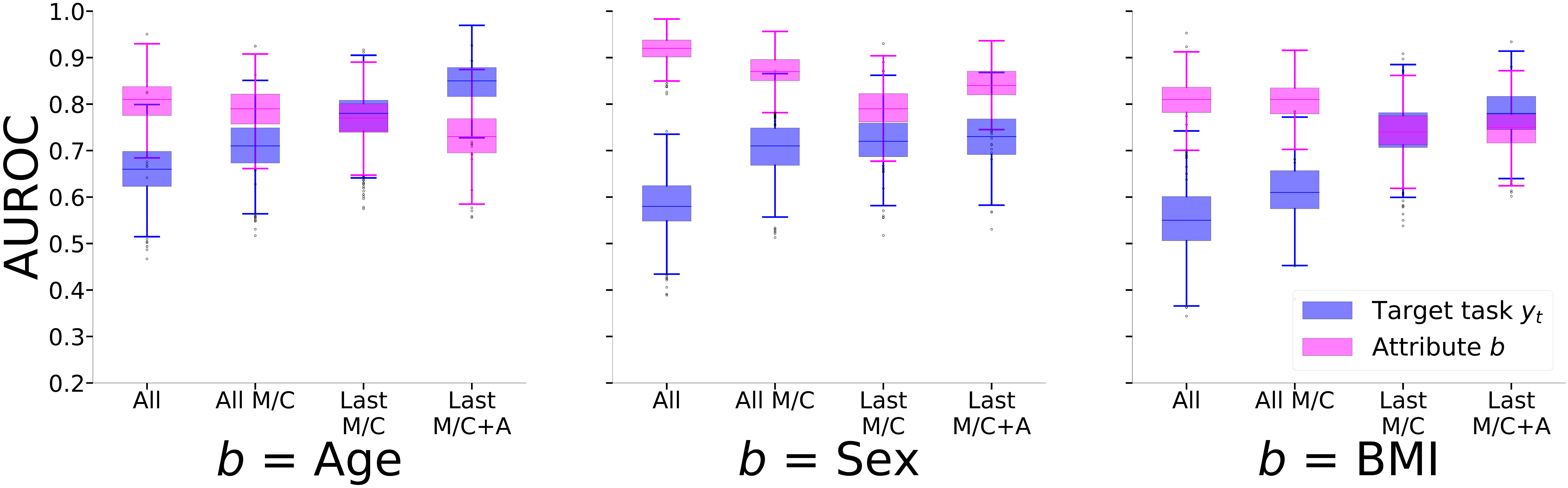}
    \caption{Across all attributes, \texttt{Last Layer M/C + A} (denoted as Last M/C + A) yields consistently higher AUROCs compared to \texttt{All Layers} (All) in predicting the target task. Moreover, \texttt{Last Layer M/C + A} performs consistently worse in terms of the predictive performance of $b$ (AUROC($\hat{y}$, $b$)) compared to \texttt{All Layers} and \texttt{All Layers M/C} (All M/C), and further drops in performance compared to \texttt{Last Layer M/C}  (Last M/C) when $b$=age. Additionally, \texttt{Last Layer M/C + A} consistently outperforms \texttt{All Layers M/C} and \texttt{Last Layer M/C} in predicting CHF.  This suggests that the proposed transfer learning approach encourages models to rely \textit{less} on shortcut features and more on clinically relevant attributes of CHF.  Somewhat surprisingly, simply applying a linear classifier to the features of a CNN allows us to predict CHF with good discriminative performance, while also mitigating, in part, the use of shortcuts.}
    \label{fig:baseline_results}
\end{figure}

Again, we measure performance in terms of AUROC on the held-out test data with respect to $y_t$ (AUROC($\hat{y}, y_t$)), which answers the question ``How well does the model predict CHF?'' In addition, for age, sex, and BMI, we compare to the predictive performance of $b$ (AUROC($\hat{y}, b$)), which tells us ``How much does the model rely on shortcuts?'' When the model takes a shortcut, we expect AUROC($\hat{y}, b$) to be high and AUROC($\hat{y}, y_t$) to be low, and vice-versa when the model does not take a shortcut. These results are shown in \textbf{Figure \ref{fig:baseline_results}}. Race and pacemaker are too infrequent in our dataset to calculate (AUROC($\hat{y}, b$)), so we report AUROC($\hat{y}, y_t$) in \textbf{Appendix Table  \ref{tab:baseline_results_table}} along with 95\% confidence intervals for all tests.\\

\par \noindent \textbf{Results and Discussion.} Across all attributes, \texttt{Last Layer M/C + A} yields consistently higher AUROCs compared to \texttt{All Layers} (\textbf{Figure \ref{fig:baseline_results}}, \textbf{Appendix Table \ref{tab:baseline_results_table}}). Using a paired resampling approach, differences in the context of all attributes appear significant (\textbf{Appendix Table \ref{tab:statistical_tests}}). Furthermore, there is a drop in the predictive performance of each attribute $b$ (i.e., AUROC($\hat{y}$, $b$)) compared to \texttt{All Layers} and \texttt{All Layers M/C}, and a further drop in performance compared to \texttt{Last Layer M/C} when $b$=age. Moreover, \texttt{Last Layer M/C + A} performs better than both \texttt{All Layers M/C} and \texttt{Last Layer M/C}, which both only use MIMIC-CXR and CheXpert for initialization. We hypothesize that, since patients with AHRF are generally sicker, there are some differences in the dataset distributions between AHRF, and MIMIC-CXR and CheXpert. Thus, retuning all layers on the AHRF source task yields better performance on the closely related AHRF target task, compared to retuning on MIMIC-CXR/CheXpert alone. These results suggest that i) features used in predicting pneumonia can also be used in predicting CHF and ii) training to learn pneumonia first, freezing the features $\thetaB$ of the model, and then using these features to predict CHF mitigates, at least in part, the use of shortcuts. We also examined the effects of tuning a varying number of denseblocks in the network (rather than just the last layer). We found that (i) model performance decreases as the number of tuned denseblocks increases, and (ii) model performance increases for some attributes $b$ when we tune all layers of the network compared to tuning three denseblocks. We suspect that, as the parameter space increases from tuning three denseblocks to tuning all layers, the model might need longer to train (\textbf{Appendix \ref{sensitivity analysis}}). Finally, we also compared to a multitask approach in which a model is trained to jointly predict $y_s$ and $y_t$, minimizing the sum of the respective cross-entropy losses. For the majority of attributes, \texttt{Last Layer M/C + A} outperforms the multitask approach, suggesting that the multitask approach, although sharing features between the unbiased source task $y_s$ and biased target task $y_t$, still allows the model to learn shortcut features (\textbf{Appendix \ref{multitask}}).

Although pneumonia oftentimes presents with more \textit{textural} features (e.g., fluid in the lungs), these features appear to still be useful in predicting CHF, which presents with more \textit{structural} features (e.g., an enlarged heart). Additionally, somewhat surprisingly, simply applying a classifier to the features of a CNN allows us to predict CHF with good discriminative performance, while also mitigating, in part, the use of shortcuts. While the size of the final linear layer is large (i.e., 1024), we do not find that it is enough to overfit to shortcut features. 




\subsection{Experiments with Preprocessing Shortcuts} 
\label{preprocessing}

Earlier we introduced skew or bias in the data by subsampling the training sets. This renders comparison across different levels of skew invalid, since other aspects of the training data could affect performance. To control for this, in this section, we explore a synthetic shortcut added through a preprocessing step. In image preprocessing, it is not uncommon to apply some amount of filtering to reduce noise. Here, a subset of the images is filtered using a Gaussian filter with standard deviation $s_1$ and the remainder are filtered with a standard deviation $s_2$. Injecting bias in this way allows us to control the correlation between the strength of the filter applied (i.e., $b$) and the target task, without changing which examples are included in the training set. 

\subsubsection{Injecting Synthetic Bias Through Preprocessing} 
\label{infer_Gaussian}

First, just as we did in \textbf{Section \ref{infer bias}}, we measured the extent to which models could learn the attributes of a Gaussian filter applied to the AHRF cohort. We defined a binary attribute $b \in \{0,1\}$ that determined which filter was applied to each image. We applied a Gaussian filter with standard deviation $s_1$ to images when $b=1$ and standard deviation $s_2$ to images when $b=0$. Given a set of $n$ chest X-rays $\{X^{(i)}, b^{(i)}\}_{i=1}^n$, we consider the task of predicting $b$, i.e., which Gaussian filter was applied to the image. 25\% of the data were filtered using $s_1$ and the remainder were filtered using $s_2$.  We varied $s_1$ and $s_2$ and measured the AUROC with respect to $b$ (\textbf{Table \ref{tab:Gaussian_filter_range_of_auc}}). Through an iterative approach, this yielded an easy, moderate and difficult task based on the AUROCs (see \textbf{Appendix \ref{filter_descriptions}} for more detail on the tasks).

\begin{table}[!ht]
\centering
\caption{On the \textit{easy shortcut} and \textit{moderate shortcut} tasks, a model can predict which Gaussian filter was applied to a chest X-ray with good discriminative performance (similar to the task of sex$=$F). However, the model struggles to differentiate between the two Gaussian filters in the \textit{difficult shortcut} task (similar to the task of insurance).}
\label{tab:Gaussian_filter_range_of_auc}

\begin{tabular}{cc}
    \toprule 
    \textbf{Ease of shortcut} &   \textbf{AUROC (95\% CI)}\\ 
    \midrule 
        \textbf{difficult shortcut}   &  0.73 (0.66 - 0.82) \\
        \textbf{moderate shortcut}  & 0.85 (0.79 - 0.91) \\ 
        \textbf{easy shortcut}  & 0.98 (0.96 - 1.00)\\ 
    \bottomrule 
\end{tabular}

\end{table}

\begin{figure}[!ht]
    \centering
       \includegraphics[width=0.5\textwidth]{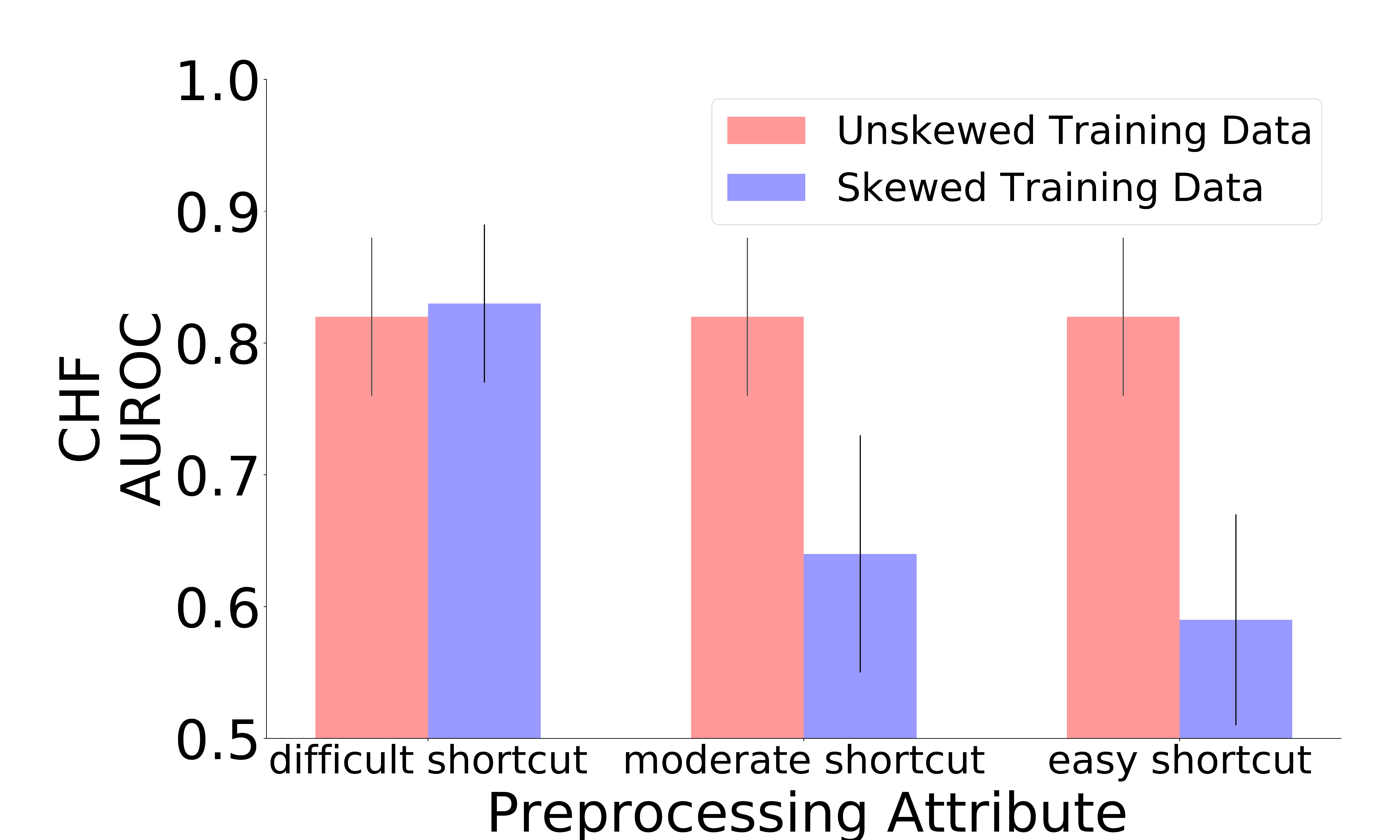}
    \caption{We compare the AUROC performance of a model trained on a synthetic dataset with no attribute skew, {\tt Unskewed Training Data}, to the performance of a model trained on a synthetic dataset with attribute skew, {\tt Skewed Training Data}. Error bars represent the 95\% CI. When the shortcut is easy or moderate, the model trained on data that closely mimics the test set significantly outperforms the model trained on data that is skewed towards the attribute.}
    \label{fig:Gaussian_skewed_vs_unskewed}
\end{figure}

Second, as in \textbf{Section \ref{shortcuts exploited}}, we introduce a 1:1 correlation between the bias and target task label (i.e., {\tt Skewed Training Data}) and compare the model's performance to a model trained on data that more closely mimics the test set (i.e., {\tt Unskewed Training Data}). As we saw previously, the presence of skew significantly affects performance when the shortcut is `easy' to learn (\textbf{Figure \ref{fig:Gaussian_skewed_vs_unskewed}, Appendix Table \ref{tab:tests_gauss_us})}.

\begin{figure}[!ht]
    \centering
    \includegraphics[width=1\textwidth]{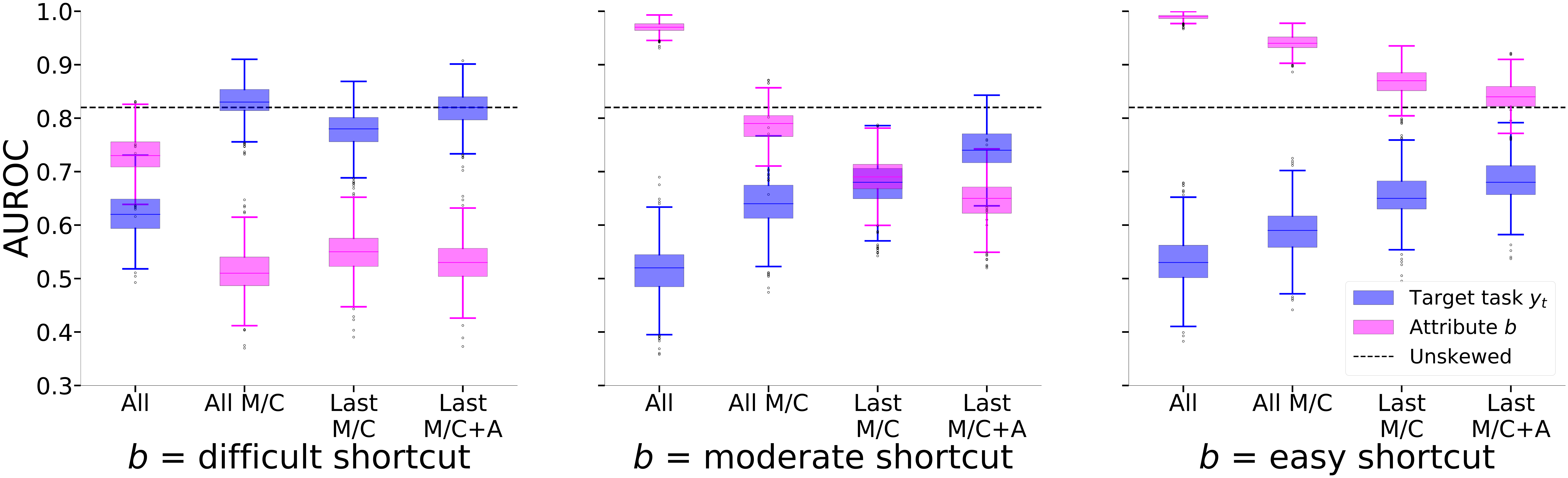}
    \caption{\texttt{Last Layer M/C + A} (denoted as Last M/C + A) outperforms \texttt{All Layers} (All) on all three synthetic datasets in predicting the target task, regardless of how difficult the shortcut is to learn. Additionally, when the shortcut is more difficult to learn, \texttt{Last Layer M/C + A} performs just as well as training on an unskewed dataset (AUROC=0.82 [95\% CI: 0.75-0.87] vs 0.82 [95\% CI: 0.76-0.88]). Moreover, when measuring the predictive performance of attribute $b$, \texttt{Last Layer M/C + A} yields consistently lower AUROCs compared to the alternative approaches when the shortcut is easy or moderate. Finally,  with the exception of when the shortcut is difficult to learn, \texttt{Last Layer M/C + A} performs better than \texttt{All Layers M/C} (All M/C) and \texttt{Last Layer M/C} (Last M/C), suggesting that retuning all layers on the AHRF source task helps.}
    \label{fig:Gaussian_filter_baselines}
\end{figure}

Third, we measured the effectiveness of the presented transfer learning approach in preventing a model from learning the attributes of a Gaussian filter, comparing to alternative approaches, just as we did in \textbf{Section \ref{2-Stage}}. Here, we aim to better understand the effectiveness of the proposed approach while controlling for the training set across models. Again, when the model takes a shortcut, we expect AUROC($\hat{y}, b$) to be high and AUROC($\hat{y}, y_t$) to be low, and vice-versa when the model does not take a shortcut. Similar to the results in \textbf{Section \ref{2-Stage}}, across all three shortcuts, \texttt{Last Layer M/C + A} (Last M/C + A) yields higher AUROCs compared to \texttt{All Layers} (All) (\textbf{Figure \ref{fig:Gaussian_filter_baselines}, \textbf{Appendix Tables \ref{tab:Gaussian_filter_baselines}}, \ref{tab:statistical_tests_synthetic}}). Moreover, \texttt{Last Layer M/C + A} performs better than both \texttt{All Layers M/C} (All M/C) and \texttt{Last Layer M/C} (Last M/C), with the exception of \texttt{All Layers M/C} for the \textit{difficult shortcut} (AUROC=0.82 [95\% CI: 0.75-0.87]) vs AUROC=0.83 [95\% CI: 0.77-0.89])). Finally, when measuring the predictive performance of attribute $b$, \texttt{Last Layer M/C + A} yields consistently lower AUROCs compared to the alternative approaches, again with the exception of \texttt{All Layers M/C} for the \textit{difficult shortcut}.

Given this setup, we explore relaxing the assumption that our source task $y_s$, is unbiased with respect to attribute $b$. We made this extreme assumption in \textbf{Section \ref{2-Stage}}, but in reality, it might not be possible to have a completely unbiased source task. Here, we explore if a transfer learning approach still applies in settings where the source task is also biased (in part).

Again, we aim to learn a mapping from the input chest X-ray $X$ to a prediction for $y_t$. However, here, we assume the source task may have some skew (i.e., $y_s$ may be correlated with $b$).  We vary the amount of correlation between pneumonia (the source task) and $b$ from $-1$ to $1$ in increments of 0.2. We hypothesize that, as the correlation increases (in either direction), the CNN will increasingly exploit the shortcut associated with the Gaussian filter when predicting pneumonia. This will then allow the target task to exploit the same shortcut, leading to poor performance on the target task test set (i.e., lower AUROC).

On held-out test data, we compare the models' discriminative performance to \texttt{All Layers}, \texttt{All Layers M/C}, and \texttt{Last Layer M/C}. As in our previous experiments, we hypothesize that pretraining on MIMIC-CXR and CheXpert, and then fine tuning the last layer of the model, will partly help mitigate the use of shortcut features. However, we hypothesize that fine tuning on the curated source task dataset will lead to better performance when the correlation between the source task and attribute is low. \\

\begin{figure}[!htbp]
    \centering
      \includegraphics[width=\textwidth]{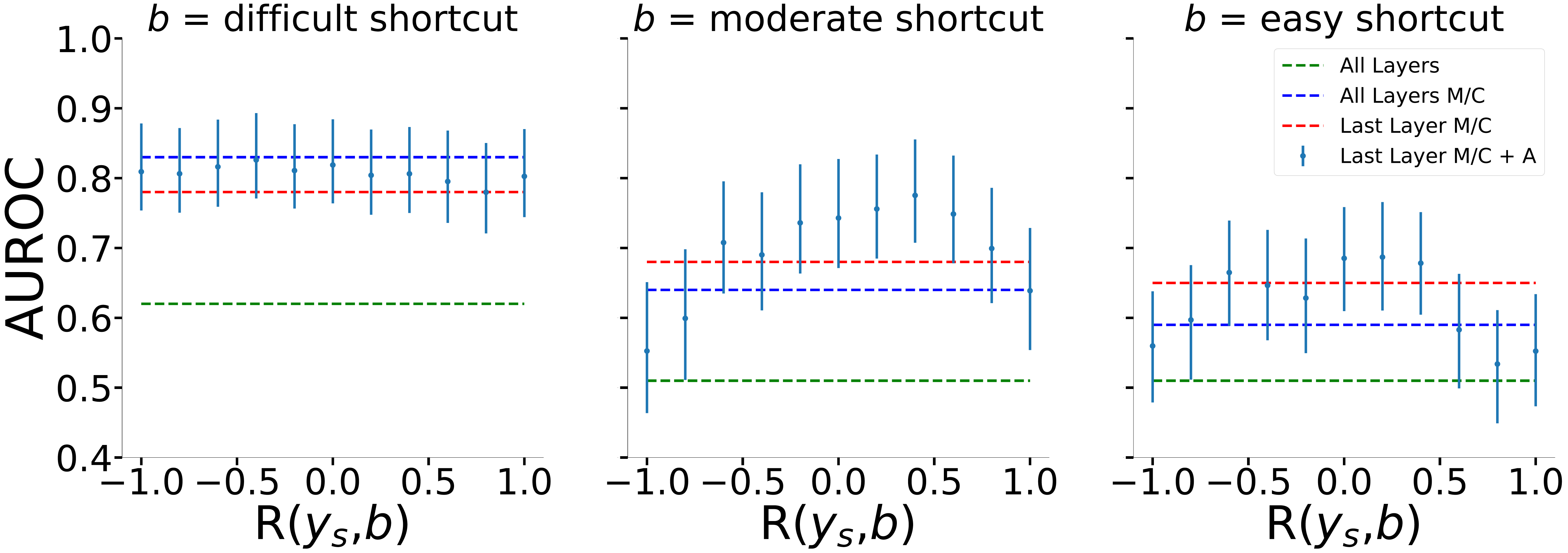}
    \caption{When we relax the assumption that the source task $y_s$ is not skewed towards the synthetic attribute $b$, \texttt{Last Layer M/C + A} consistently outperforms \texttt{All Layers}. Error bars represent the 95\% CI. Additionally, when small amounts of skew are introduced, for the \textit{easy shortcut} and \textit{moderate shortcut} datasets, the model outperforms both \texttt{All Layers} and \texttt{Last Layer M/C} . However, when the shortcuts are more difficult to learn (\textit{difficult shortcut}), we do not see improvements over \texttt{All Layers M/C}, and only marginal improvements over \texttt{Last Layer M/C}. Finally, when \textit{too} much skew is introduced, model performance for both the \textit{moderate shortcut} and \textit{easy shortcut} degrades close to that of \texttt{All Layers}.}
    \label{fig:pna_correl}
\end{figure}

\par \noindent \textbf{Results and Discussion.} When we relax the assumption that the source task exhibits no correlation between the label $y_s$ and attribute $b$, \texttt{Last Layer M/C + A} performs as well or better than \texttt{All Layers}. (\textbf{Figure \ref{fig:pna_correl}}). Additionally, when small amounts of skew are introduced, in the \textit{easy shortcut} and \textit{moderate shortcut} setting, \texttt{Last Layer M/C + A} outperforms both \texttt{All Layers M/C} and \texttt{Last Layer M/C}, again emphasizing the importance of the domain-specific knowledge contained in the curated source task dataset. When the shortcuts are difficult to learn, all of the models with the exception of \texttt{All Layers} perform similarly, regardless of the correlation between the pneumonia label and $b$. Finally, when a larger amount of skew is introduced between the pneumonia label and $b$ (e.g., a correlation closer to -1 or 1), model performance for both the \textit{moderate shortcut} and \textit{easy shortcut} decreases significantly.

Overall, these results suggest that we can \textit{partly} relax the assumption that our source task $y_s$ is not correlated with attribute $b$. This may explain why pretraining or transfer learning with the MIMIC-CXR/CheXpert data offers some improvement over \texttt{All Layers}, despite no guarantee that the data do not exhibit any skew.


%% file: mlhc-submission-files/text/discussion.tex
\section{Conclusion} 

This work explores the extent to which deep nets can exploit undesirable `shortcuts' in diagnostic tasks involving chest X-rays and potential solutions for mitigating such behavior. We demonstrated that deep learning models can learn demographic features such as age and BMI, as well as subtle differences in image preprocessing steps, from chest X-rays with high discriminative performance. Additionally, we showed that in the presence of spurious correlations, deep learning models can exploit potential shortcuts associated with biased attributes rather than learning clinically relevant features, leading to poor generalizability. We investigated a simple transfer learning approach based on pretraining and fine-tuning only the output layer of a DenseNet to mitigate the use of these shortcuts. We showed that this approach could, in part, produce models with improved generalization performance despite training on skewed data.

We note some important limitations of our work. First, we only considered a subset of attributes for which we had data. The number of attributes that can lead to skew in datasets can go far beyond what is presented in this work. Additionally, we considered these attributes separately in each of our experiments. It would be beneficial to test the ability of the proposed approach to mitigate the effect of multiple sources of skew. Finally, this approach relies on the idea that features for one diagnostic task can be reused for another diagnostic task. However, not all diagnoses necessarily share similar features. Despite these limitations, our results suggest that the transportability of features in diagnostic tasks has the potential to mitigate the use of shortcut features in the presence of bias, leading to better generalizability across populations.